\documentclass{fairmeta}
\usepackage{amsthm}

\usepackage{amsmath,amsfonts,bm}

\def\eqref#1{equation~\ref{#1}}

\def\1{\bm{1}}

\DeclareMathAlphabet{\mathsfit}{\encodingdefault}{\sfdefault}{m}{sl}
\SetMathAlphabet{\mathsfit}{bold}{\encodingdefault}{\sfdefault}{bx}{n}

\usepackage{hyperref}
\usepackage{url}
\usepackage{graphicx}
\usepackage{array}
\usepackage{booktabs}
\usepackage{tabularx}
\usepackage{longtable}
\usepackage{multirow}
\usepackage{float}
\usepackage{fvextra}

\title{MLToolBench: Learning Tool-Augmented Agents for Machine Learning Development}

\author[1,*]{Xin Yu}
\author[2,*]{Lizhu Zhang}
\author[1,*]{Jiamu Bai}
\author[2]{Yanhong Wu}
\author[2]{Zellux Wang}
\author[2]{Serena Li}
\author[2]{WEIWEI LI}
\author[1]{Lingzhou Xue}
\author[2]{Xiangjun Fan}
\author[2,\dagger]{Bo Peng}
\contribution[*]{Equal contribution (co-first authors)}
\contribution[\dagger]{Corresponding author}
\hypersetup{pdfauthor={Xin Yu, Lizhu Zhang, Jiamu Bai, Yanhong Wu, Zellux Wang, Serena Li, WEIWEI LI, Lingzhou Xue, Xiangjun Fan, Bo Peng}}

\newcommand{\toolbench}{MLToolBench}
\newcommand{\spice}{\textsc{Spice}}
\newcolumntype{L}[1]{>{\raggedright\arraybackslash}p{#1}}
\newcolumntype{Y}{>{\raggedright\arraybackslash}X}

\abstract{Machine learning engineering (MLE) agents have made substantial progress, but learning through ML experimentation remains costly in time and computation. Synthetic environments reduce these costs while introducing variations in data and experimental settings that require task-specific diagnosis. Access to diagnostic tools alone does not ensure that agents learn when to use them or how to act on their findings. We introduce \toolbench{}, a suite of executable tools for data inspection, code verification, and experiment diagnosis, together with an SFT and RL pipeline for learning their use. Diagnostic calls acquire evidence whose value depends on subsequent decisions, so final outcomes provide limited guidance on which calls to reinforce. We address this challenge with \spice{}, which measures how privileged context changes the likelihood of a sampled tool action and uses this difference as a turn-level reward alongside the final outcome. We train on 80 synthetic tasks and evaluate on 25 in-domain and 10 out-of-domain tasks. Providing tool interfaces and descriptions alone yields inconsistent gains across unadapted models. With the same diagnostic interface, our training pipeline raises in-domain success from 24.8\% to 52.4\% for Qwen3-8B and from 35.6\% to 69.2\% for Qwen3.5-35B-A3B. The latter also improves from 31\% to 48\% out-of-domain, supporting learned diagnostic tool use on held-out sources and targets.}
\date{November 26, 2026}
\hypersetup{pdftitle={MLToolBench: Learning Tool-Augmented Agents for Machine Learning Development}}
\begin{document}
\maketitle

\begin{figure}[!ht]
\centering
\includegraphics[width=\textwidth]{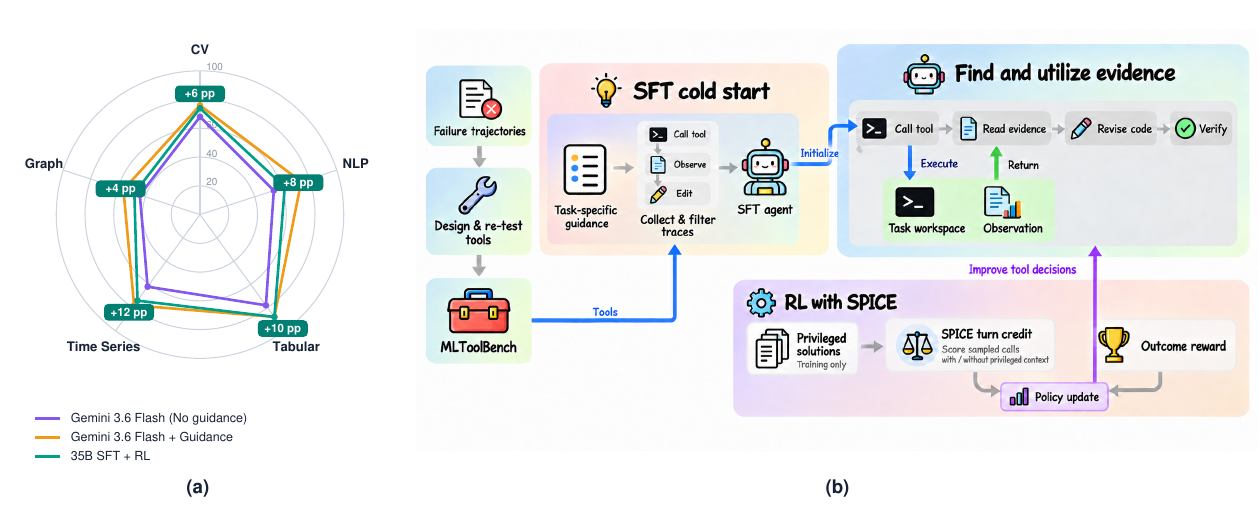}
\caption{\textbf{Learning to find and utilize diagnostic evidence.} \textbf{(a)} In-domain success under three setups: Gemini 3.6 Flash with and without guidance, and Qwen3.5-35B-A3B after SFT + RL, all using \toolbench{}. Guided Gemini receives solution-informed hints. Signed pp labels show Qwen's gains over unguided Gemini. \textbf{(b)} Failure-driven tool construction, guided SFT, and RL with \spice{} teach agents to acquire and use diagnostic evidence.}
\label{fig:overview}
\end{figure}

\section{Introduction}

Machine learning engineering (MLE) agents increasingly perform multi-step ML development tasks \citep{liu2025mlagent,yang2026rlmle}. Learning from these interactions is costly. Verifying a single rollout may require preprocessing data, training a model, and evaluating its output. Repeating this process places substantial demands on time and computation \citep{cai2026acegrpo}. Recent work develops verifiable tasks and synthetic environments for MLE agents, with small synthetic datasets reducing the cost of execution and verification during RL rollouts \citep{qiang2025mlesmith,zhou2026sandmle}. Synthetic task generation can also be combined with pass-rate-based task selection to support RL curricula that provide challenges suited to the model's current capabilities \citep{mai_thinking_1}.

\begin{figure}[t]
\centering
\includegraphics[width=\textwidth]{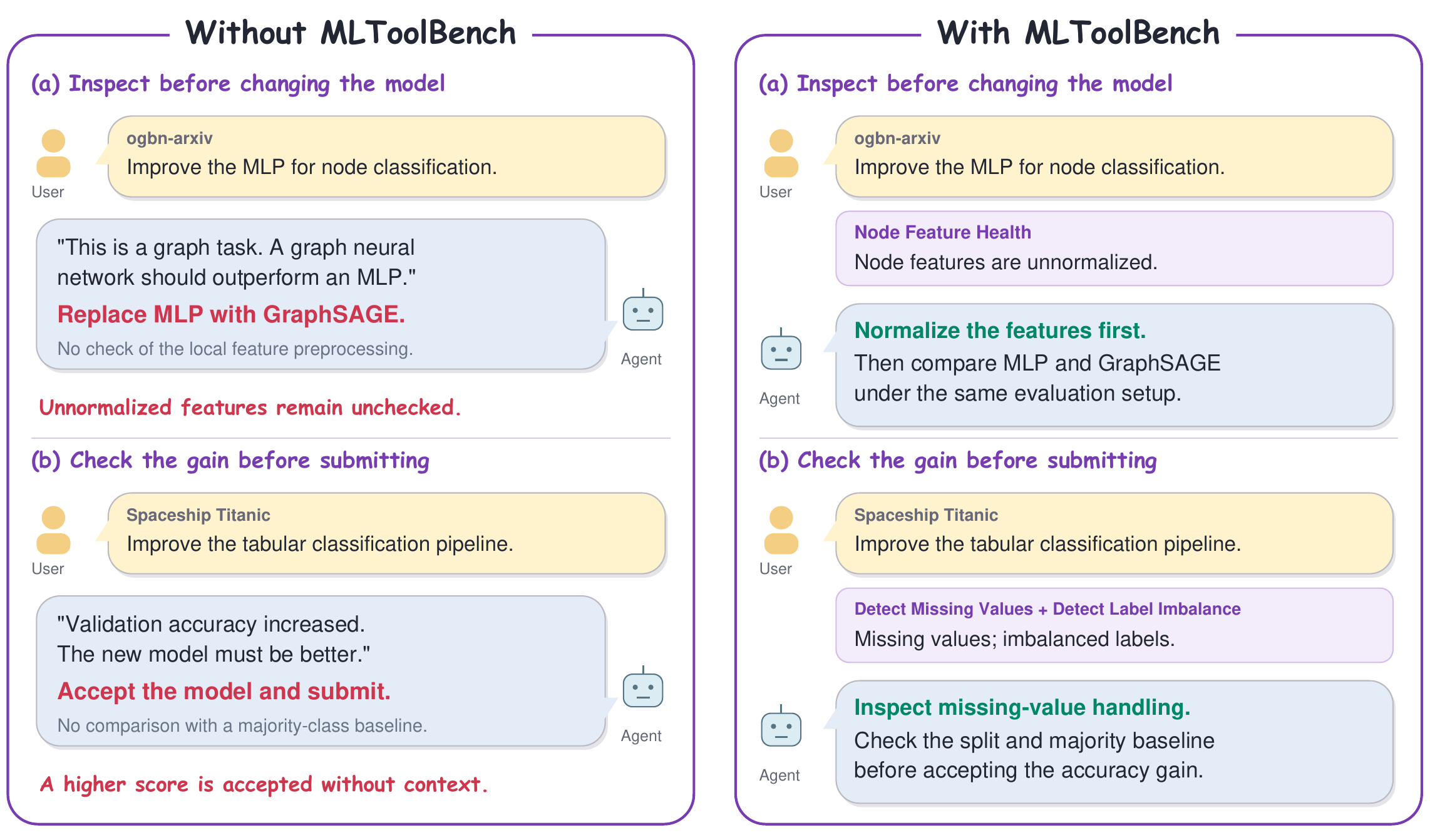}
\caption{Schematic motivating cases for \toolbench{}. \textbf{Top:} local feature diagnostics guide preprocessing checks before an architectural change. \textbf{Bottom:} missingness and imbalance motivate checking an apparent validation gain before submission. The contrast illustrates why evidence acquisition should precede intervention; Section~\ref{sec:experiments} separately evaluates aggregate task performance.}
\label{fig:motivating_cases}
\end{figure}

These environments make interaction more affordable, but agents must still diagnose the task they actually encounter. A synthetic instance may resemble a familiar ML problem while changing its data distribution, preprocessing, or evaluation conditions. Two failure patterns illustrate this challenge. First, \textbf{mismatched modeling priors} arise when agents select code modifications based on familiar strategies without inspecting the local data. In Figure~\ref{fig:motivating_cases} (top), the agent replaces an MLP with GraphSAGE before checking whether node features are normalized. Second, \textbf{insufficient baseline comparison} occurs when agents treat a validation score as evidence of improvement without comparing it against a simple reference model. In Figure~\ref{fig:motivating_cases} (bottom), class imbalance makes comparison with a majority-class baseline essential before accepting the apparent gain. These concerns extend beyond synthetic environments. Real-world MLE tasks span diverse datasets and experimental requirements, making it difficult to anticipate every check an agent will need. MLAgentBench reports agents discarding features before assessing their predictive value \citep{huang2024mlagentbench}, while MLE-bench documents failures to use available submission checks, resulting in invalid outputs \citep{chan2025mlebench}. Across these settings, the common challenge is to connect task-specific evidence to the next decision. Execution tools make checks possible, but agents must learn which checks to perform and how to use their results. We therefore ask: \emph{How can MLE agents learn to find and utilize diagnostic evidence effectively?}

In this work, we introduce \toolbench{} and a training framework that jointly address this capability gap. Built on the MLAgentBench scaffold, \toolbench{} exposes data inspection, code verification, and experiment diagnosis as explicit actions, with tools derived from recurring agent failures. Agents acquire evidence by invoking these tools; their interfaces and descriptions alone provide no practical experience in interpreting the results. Guided SFT establishes inspection-to-action behavior, and RL refines it through task interaction. The tool library and learning method thus serve the same objective: learning to obtain diagnostic evidence and use it to guide ML development.

Diagnostic tool use creates a particular credit-assignment challenge. Its benefit often emerges through a later decision, such as normalizing features after inspecting their scale. Final task rewards do not identify which earlier inspections informed that decision. Potential-based methods such as TIPS track changes in reference-answer likelihood \citep{xie2026tips}, yet useful diagnostic evidence need not immediately increase that likelihood. On fixed SFT trajectories, we observe smaller mean absolute TIPS scores for diagnostic calls than for editing/execution actions across three fixed policy checkpoints (Figure~\ref{fig:credit_by_action_type}). Motivated by this distinction, we propose \emph{Solution-Privileged Information Credit Estimation} (\spice{}), inspired by knowledge distillation with privileged information \citep{lopezpaz2016gdistill}. A frozen snapshot of the updated policy model scores each sampled action with and without a verified solution summary. The resulting log-likelihood difference measures contextual support for the action and supplies signed turn-level feedback alongside the outcome reward. This gives diagnostic decisions a learning signal without requiring an immediate increase in reference-answer likelihood.

Our contributions are as follows:
\begin{itemize}
\item \textbf{An executable diagnostic interface for MLE.} We construct \toolbench{} from recurring agent failures, exposing data, code, and experiment checks as explicit actions within MLAgentBench. Task reruns verify candidate tools before their inclusion in the library.
\item \textbf{A training framework for diagnostic decisions.} Guided SFT establishes initial tool-use behavior. RL then combines task outcomes with SPICE, which scores sampled actions using the change in their likelihood under privileged solution context.
\item \textbf{Evidence for learned diagnostic tool use.} Across held-out ID and OOD tasks, the full pipeline improves both Qwen backbones. Interface comparisons, matched RL controls, and diagnostic-observation interventions examine tool access, training, and evidence use separately.
\end{itemize}

\section{Method}
\label{sec:method}

\noindent\textbf{Method overview.} Our pipeline connects tool construction, demonstration learning, and reinforcement learning (Figure~\ref{fig:overview}b). MLToolBench makes diagnostic evidence accessible through executable calls. Guided SFT teaches an initial pattern of inspection and intervention; RL refines the policy using task outcomes and SPICE turn-level feedback. The trained student acquires observations from the workspace and receives no privileged solution context at evaluation.

\noindent\textbf{Notation.} A task $x$ contains an initial workspace, instruction, and executable evaluator; $\mathcal T_x$ denotes its workspace and diagnostic actions. At interaction index $s$, policy $\pi_\theta$ generates an assistant action $a_s$ from history $h_{s-1}$ and receives an observation $o_s$. We call $z_s=(a_s,o_s)$ a segment and write $h_s=(x,\mathcal T_x,z_{1:s})$ for the resulting history. A rollout is $\tau=(z_1,\ldots,z_S)$. We use $t$ for a token position within assistant-generated text, with token $y_t$ and ordinary prefix $H_t$. The evaluator assigns an outcome $R(\tau)$ to the rollout; \spice{} scores sampled tool calls to construct turn-level rewards. SFT and RL optimize the same parameters $\theta$.

\subsection{MLToolBench: Diagnostic Interface Construction}
\label{sec:toolbench}

\begin{figure}[!htbp]
\centering
\includegraphics[width=\textwidth]{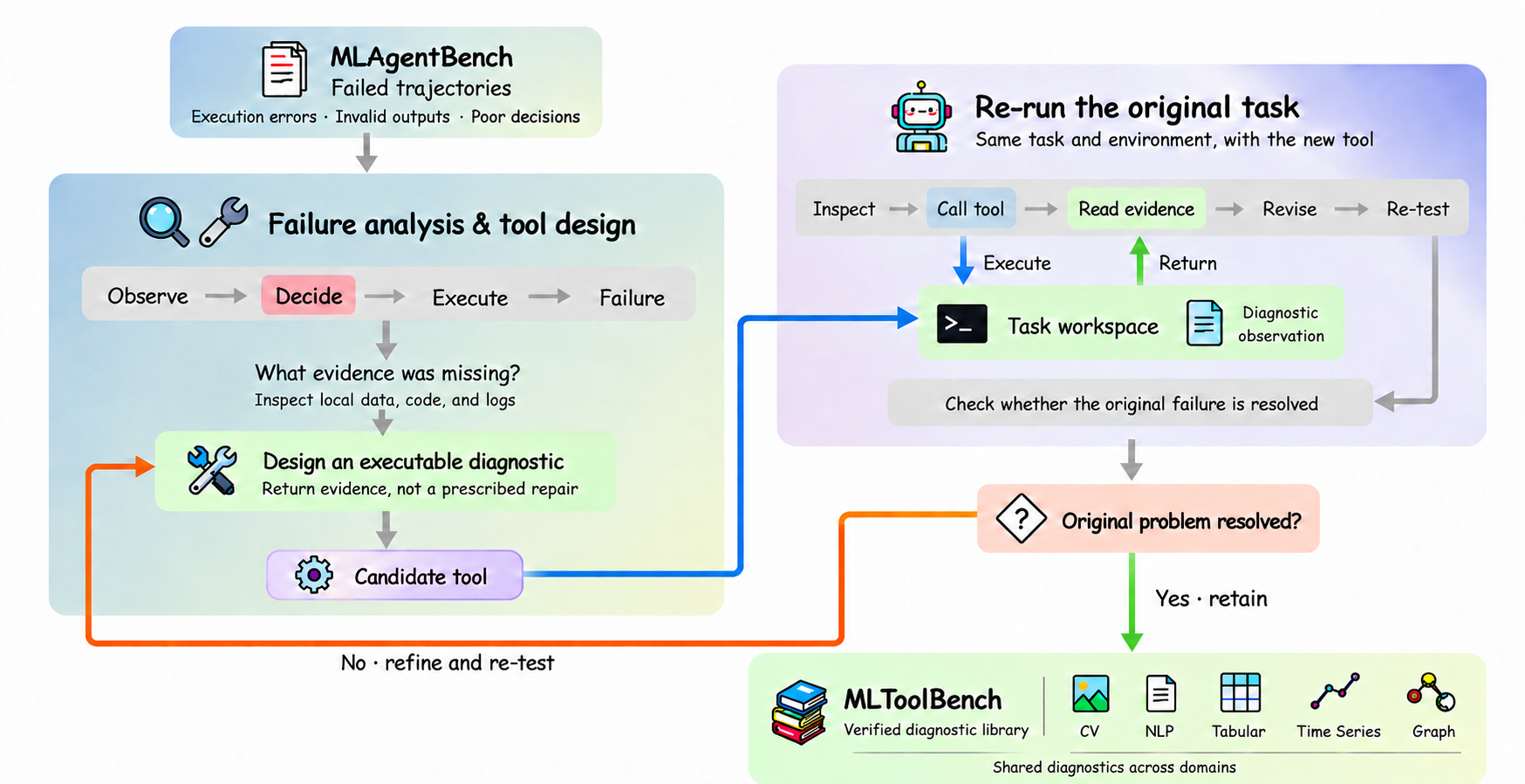}
\caption{\textbf{Construction of \toolbench{}.} Failure analysis motivates diagnostic tool candidates. The agent re-runs the original task with each candidate; tools that help resolve the original failure are retained and organized by domain, while unsuccessful candidates are refined and re-tested.}
\label{fig:toolbench_construction}
\end{figure}

Our benchmark pairs a diagnostic tool library with executable ML task instances for training and evaluation. \toolbench{} extends MLAgentBench's workspace interface with 67 registered diagnostic tools: seven shared tools and 60 tools organized into tabular, time-series, NLP, vision, and graph modules \citep{huang2024mlagentbench}. They support three functions: \emph{data inspection} (e.g., missing values, label balance, and node-feature health), \emph{code verification} (e.g., configuration consistency, tensor shapes, and output formats), and \emph{experiment diagnosis} (e.g., metric extraction, trends, and plateaus). Time-series tools also analyze training curves across domains. Figure~\ref{fig:toolbench_construction} summarizes construction: we identify missing evidence in failed trajectories, design a diagnostic, and re-run the original task with the candidate tool. Candidates are retained when the original failure is resolved; otherwise, they are refined and re-tested. Agents call these tools to obtain measurements and, in some cases, heuristic suggestions, then choose subsequent edits themselves. Appendix~\ref{app:toolbench_details} lists the complete library and distinguishes it from each run's exposed tools.

\subsection{Tool-Use Cold Start via Supervised Fine-Tuning}
\label{sec:guided_sft}

\noindent\textbf{Constructing task-specific guidance.} We use an LLM to assess each training task from its specification, reference solution, and synthetic environment generation mechanism. It tailors guidance to the instance's prediction target, data characteristics, starter pipeline, and evaluation constraints, specifying which diagnostic tools to call and how alternative observations should inform subsequent actions. The guidance provides an investigation strategy rather than a target patch; the agent must still invoke tools to obtain evidence from the workspace.

\noindent\textbf{Collecting demonstrations for a tool-use cold start.} We insert this guidance into Gemini 3.6 Flash's prompt alongside the task and available tool descriptions. The model then interacts with the environment to produce demonstrations that connect tool calls and observed evidence to subsequent actions. We filter these trajectories and use them for SFT, removing the guidance from student inputs while retaining the actual interactions. Standard next-token supervision is applied only to assistant-generated tokens; tool observations remain context. This gives the policy an initial ability to use diagnostic tools before RL. Appendix~\ref{app:sft_details} provides collection, filtering, and training details.

\subsection{Reinforcement Learning with Turn-Level Credit}
\label{sec:rl}

After SFT, the policy learns from fresh interactions without collection-time guidance. Final task rewards evaluate the resulting ML artifact, but provide limited feedback on diagnostic decisions along the way. Checking feature scales, for example, acquires evidence whose value depends on a later preprocessing choice. We therefore propose \emph{Solution-Privileged Information Credit Estimation} (\spice{}). Inspired by knowledge distillation with privileged information \citep{lopezpaz2016gdistill}, SPICE uses solution summaries as additional scoring context. The policy acts without these summaries. Their effect on the likelihood of its sampled action supplies a turn-level reward.

\noindent\textbf{Scoring the current action.} TIPS asks whether a tool interaction makes the reference answer more likely \citep{xie2026tips}. In our ML adaptation, the reference is a multi-step solution summary. A check of feature scales may inform just one preprocessing decision, so its effect on a few reference tokens can be small when averaged over the whole summary. SPICE asks \textbf{how knowing the solution changes the probability of the current tool action}. It holds the pre-action history fixed and averages over the tokens of that action, whether it is a diagnostic call, a code edit, or an execution step. Appendix~\ref{app:spice_tips_discussion} gives the two scores. Appendices~\ref{app:diagnostic_usefulness} and~\ref{app:score_masking} test what higher scores mean for diagnostic calls.

\noindent\textbf{Scoring tool-use turns.}
\label{sec:dynamic_credit}
Each tool-use turn contains an assistant tool call and the environment's returned observation. For a sampled call $a_{i,s}$ in rollout $i$, turn $s$, we compare its likelihood under the same frozen snapshot $q$ of the updated policy model, with and without verified-solution context. Let $h_{i,s-1}$ be the pre-action history and $\mathcal C_x$ a frozen set of $K$ agent-written summaries of verified candidate solutions for the current training task. The turn score is
\begin{equation}
d_{i,s}=\frac{1}{K|a_{i,s}|}\sum_{c\in\mathcal C_x}\sum_{u=1}^{|a_{i,s}|}
\log\frac{q(a_{i,s,u}\mid h_{i,s-1},c,a_{i,s,<u})}
{q(a_{i,s,u}\mid h_{i,s-1},a_{i,s,<u})}.
\label{eq:solution_score}
\end{equation}
Both branches teacher-force the same sampled tool-call tokens. The current tool observation and future trajectory are excluded from scoring; earlier observations remain in the history. Positive and negative scores indicate increased and decreased support under privileged context. The solution bank starts from verified demonstrations and admits newly verified RL solutions between updates. Each call is scored in its own history, so trajectories need not share tools or turn counts.

\noindent\textbf{Adding turn rewards to outcome feedback.}
\label{sec:segment_grpo}
We use binary terminal success $R_i$ under Equation~\ref{eq:extended_success}, extending MLAgentBench's threshold \citep{huang2024mlagentbench} to signed and near-zero baselines with an absolute-gain floor of 0.01. The task test script evaluates submissions and starter baselines on the same partition; invalid submissions receive zero. We add the detached tool score at its turn boundary and $R_i$ at termination. With $S_i$ turns and shaping coefficient $\beta\geq0$, the reward and undiscounted return-to-go are
\begin{equation}
r_{i,s}=\mathbf 1[s=S_i]R_i+\beta\,\operatorname{sg}(d_{i,s}),
\qquad
G_{i,s}=\sum_{j=s}^{S_i}r_{i,j}.
\label{eq:credit_training}
\end{equation}
Non-tool turns and calls without a valid reference score receive zero shaping reward. To control shaping magnitude, we calibrate $\beta$ with a short training-task pilot for each backbone to target a mean absolute turn-level shaping reward in $[0.05,0.3]$ on pilot rollouts (Appendix~\ref{app:shaping_calibration}).

\noindent\textbf{Advantage estimation and policy update.} We adapt GRPO \citep{shao2024deepseekmath} to turn-level returns. For $B$ rollouts of task $x$, we center each turn's return using the mean complete return:
\begin{equation}
\mu_x=\frac1B\sum_{i=1}^{B}G_{i,1},\qquad
\widehat A_{i,s}=G_{i,s}-\mu_x.
\label{eq:outcome_advantage}
\end{equation}
We omit group standard-deviation scaling to avoid the difficulty-dependent task weighting identified by \citet{liu2025drgrpo}. Let $\mathcal P_i$ contain assistant tokens, $s(i,t)$ identify token $y_{i,t}$'s turn, and $\rho_{i,t}=\pi_\theta(y_{i,t}\mid H_{i,t})/\pi_{\mathrm{old}}(y_{i,t}\mid H_{i,t})$, where $H_{i,t}$ is its original policy context. We minimize the clipped objective
\begin{equation}
\mathcal L=-\mathbb E\left[\frac1B\sum_{i=1}^{B}\frac1{|\mathcal P_i|}
\sum_{t\in\mathcal P_i}\min\left\{
\rho_{i,t}\widehat A_{i,s(i,t)},
\operatorname{clip}(\rho_{i,t},1-\eta,1+\eta)\widehat A_{i,s(i,t)}
\right\}\right].
\label{eq:segment_grpo_loss}
\end{equation}
Here $\eta$ is the clipping threshold. Scores and advantages are detached; tool observations are context, not prediction targets. Setting $\beta=0$ recovers outcome-only training with mean-centered advantages. Appendix~\ref{app:dynamic_credit} specifies token assignment and reference scoring.

\section{Experiments}
\label{sec:experiments}

We ask four questions: (Q1) Does learning convert diagnostic access into task success? (Q2) Do gains transfer to held-out sources and targets, and does diagnostic evidence affect success? (Q3) How do training and credit assignment change tool-use behavior and feedback? (Q4) What do guided demonstrations, RL, and SPICE each contribute?

\subsection{Experimental Setup}
\label{sec:experimental_setup}

\noindent\textbf{Benchmark and task splits.} We use 80 training, 25 ID, and 10 OOD instances across vision, NLP, tabular learning, time series, and graph learning. ID reuses training templates with held-out data or generated samples; OOD holds out data sources or prediction targets within these domains. Multiple instances can derive from one source \citep{qiang2025mlesmith,zhou2026sandmle}. Evaluation instances are excluded from SFT, training solution banks, and RL. Appendix~\ref{app:working_manifest} defines all instances and splits.

\noindent\textbf{Evaluation protocol.} We report success rates using a signed-baseline extension of MLAgentBench's improvement threshold \citep{huang2024mlagentbench}: valid submissions must achieve a direction-adjusted improvement exceeding both 10\% of the baseline magnitude and an absolute-gain floor of 0.01 (Appendix~\ref{app:evaluation_metrics}). Failed or invalid runs are unsuccessful. For each fixed model or checkpoint, we run ten evaluation rollouts per task, yielding 250 ID and 100 OOD runs per condition; no retraining occurs between these repeats. \emph{Original} denotes the MLAgentBench scaffold with its standard tools for file editing and code execution, without MLToolBench diagnostics; \emph{+ MLToolBench} adds these diagnostics. These labels specify evaluation-time tool access, irrespective of training condition. Tasks and execution budgets are held fixed.

\noindent\textbf{Compared methods.} Table~\ref{tab:main_results} compares closed-source Gemini 3.6 Flash and Claude Sonnet 5 \citep{anthropic2026sonnet5} with open-source coding and Qwen baselines, including DeepSeek-Coder-33B-Instruct \citep{guo2024deepseekcoder}, under the same scaffold. Unadapted baselines receive tool interfaces and descriptions, but no demonstrations, task-specific guidance, or prior training on this interface; they can call tools and observe outputs during evaluation. Both Qwen backbones receive full SFT and RL with \spice{}; full-SFT controls without MLToolBench demonstrations appear in the ablation study (Appendix Figure~\ref{fig:sft_ablations}). Learned policies receive no collection-time guidance at evaluation.

\noindent\textbf{Training and objective controls.} Each SFT condition collects 100 Gemini trajectories per training task (8,000 before filtering). RL continues tool-enabled full SFT for 100 updates with fresh on-policy rollouts; we evaluate the final checkpoint. Equation~\ref{eq:credit_training} combines outcome and signed turn rewards. For each backbone, outcome-only and TIPS controls share the SFT initialization, training data, tool interface, rollout budget, optimization settings, and binary outcome reward with \spice{}. Outcome-only sets $\beta=0$; TIPS replaces the shaping method. Appendix~\ref{app:experimental_protocol} describes the evaluation protocol and training configuration; Appendices~\ref{app:sft_details} and~\ref{app:agent_prompts} detail SFT and prompts.

\subsection{Main Results}
\label{sec:main_results}

\begin{table}[t]
\centering
\small
\renewcommand{\arraystretch}{1.12}
\setlength{\tabcolsep}{4pt}
\begin{tabular*}{\textwidth}{@{\extracolsep{\fill}}llcccc@{}}
\toprule
 & & \multicolumn{2}{c}{In-domain} & \multicolumn{2}{c}{Out-of-domain} \\
\cmidrule(lr){3-4}\cmidrule(lr){5-6}
Backbone & Model / training & Original & + \toolbench{} & Original & + \toolbench{} \\
\midrule
\multicolumn{6}{c}{\textit{Closed-Source Models}} \\
\midrule
\multicolumn{2}{l}{Gemini 3.6 Flash} & \textbf{65.20\%} & 61.20\% & \textbf{52.00\%} & 46.00\% \\
\multicolumn{2}{l}{Claude Sonnet 5} & 64.80\% & 62.40\% & 45.00\% & 43.00\% \\
\midrule
\multicolumn{6}{c}{\textit{Open-Source Models}} \\
\midrule
\multicolumn{2}{l}{Qwen2.5-Coder-14B} & 45.20\% & 40.40\% & 34.00\% & 32.00\% \\
\multicolumn{2}{l}{DeepSeek-Coder-33B-Instruct} & 57.20\% & 49.20\% & 36.00\% & 34.00\% \\
\multicolumn{2}{l}{Qwen3-8B} & 24.40\% & 24.80\% & 23.00\% & 20.00\% \\
\multicolumn{2}{l}{Qwen3.5-35B-A3B} & 34.40\% & 35.60\% & 29.00\% & 31.00\% \\
\midrule
\multicolumn{6}{c}{\textit{Open-Source Reasoning Models (Fine-Tuned)}} \\
\midrule
\multirow{2}{*}{Qwen3-8B} & Full SFT & 36.80\% & 38.40\% & 28.00\% & 32.00\% \\
 & SFT + RL (\spice{}) & 49.20\% & 52.40\% & 28.00\% & 36.00\% \\
\cmidrule(lr){1-6}
\multirow{2}{*}{\shortstack[l]{Qwen3.5-\\35B-A3B}} & Full SFT & 39.60\% & 44.40\% & 34.00\% & 41.00\% \\
 & SFT + RL (\spice{}) & 51.20\% & \textbf{69.20\%} & 41.00\% & \textbf{48.00\%} \\
\bottomrule
\end{tabular*}
\caption{Main success rates (\%; 250 ID and 100 OOD runs per condition). Fine-tuned models use MLToolBench during training. Original uses standard MLAgentBench tools without MLToolBench diagnostics; + MLToolBench adds these diagnostics at evaluation. Bold marks column maxima.}
\label{tab:main_results}
\end{table}

\noindent\textbf{Training converts diagnostic access into task success.} With \toolbench{}, SFT followed by RL raises ID success from 24.8\% to 52.4\% for Qwen3-8B and from 35.6\% to 69.2\% for Qwen3.5-35B-A3B (Table~\ref{tab:main_results}, Q1). The 35B policy exceeds augmented-interface Claude and unguided Gemini by 6.8 and 8 points, but remains below privileged guided Gemini (74.4\%). These 27.6/33.6-point gains demonstrate policy improvement under an unchanged diagnostic interface.

\noindent\textbf{Tool descriptions alone do not ensure effective use.} Without tool-specific training or guidance, diagnostic access reduces ID success for Gemini, Claude, and both coding baselines; unadapted Qwen3-8B and Qwen3.5-35B-A3B gain only 0.4/1.2 points. For the SFT + RL policies, retaining diagnostics at evaluation raises ID success by 3.2/18 points for 8B/35B. Original measures interface removal after tool-enabled training; Section~\ref{sec:ablations} separately evaluates SFT without MLToolBench. This contrast motivates learning to use diagnostics beyond exposing their interfaces.

\subsection{Generalization and Diagnostic Utility}
\label{sec:generalization}

\noindent\textbf{Generalization.} On held-out sources and prediction targets, SFT followed by RL raises OOD success with \toolbench{} from 20\% to 36\% for 8B and from 31\% to 48\% for 35B (Table~\ref{tab:main_results}). Diagnostic access improves the trained policies by 8 and 7 points, respectively. The 35B policy attains 48\%, compared with Claude at 43\% and Gemini at 46\% under the same augmented interface. These tasks use held-out sources or targets within our five training domains.

\noindent\textbf{SPICE gains carry over to external tasks and new tools.} On the 30 official MLS-Bench-Lite tasks, the frozen 35B SPICE checkpoint scores \textbf{24.6}, compared with 19.2 for TIPS, 18.1 for outcome-only RL, 16.2 for SFT, and 12.8 for the original checkpoint. We use each checkpoint without further training (Appendix~\ref{app:external_mls}). On five tabular tasks, giving SPICE access to Evidently tools it had never used during SFT or RL raises success from 48\% to 66\% (Appendix~\ref{app:unseen_tools}).

\noindent\textbf{Masking diagnostic results reduces SPICE's success.} Across 250 ID rollouts per policy, masking returned diagnostic information lowers success by 1.6 percentage points for outcome-only RL and 14.8 for \spice{}. Paraphrasing the same returns lowers success by 0.4 and 0.8 points, respectively. For SPICE, this amounts to 37 fewer successes with masking and two fewer with paraphrasing (Table~\ref{tab:observation_intervention}). Both interventions leave tool access intact. Each policy chooses its own calls, so we compare each with its own unmodified returns. Appendix~\ref{app:score_masking} separates high- and low-scoring calls.

\subsection{Credit Assignment and Tool-Use Behavior}
\label{sec:behavioral_analysis}

\noindent\textbf{Training broadens diagnostic tool use.} For Q3, 35B data-management/verification coverage rises from 0.3/0.5 to 2.7/4.7 rollouts out of five (Appendix Figure~\ref{fig:tool_coverage}). The 8B RL policy achieves higher success with similar coverage but more distinct tools (six versus two). For 35B, SPICE uses eight distinct MLToolBench tools in 11 calls on average, versus two in nine calls for outcome-only RL (Appendix Table~\ref{tab:tool_use_statistics}). TIPS averages three tools in ten calls. Appendix~\ref{app:diagnostic_usefulness} evaluates the usefulness of individual calls.

\begin{figure}[!htbp]
\centering
\includegraphics[width=\linewidth]{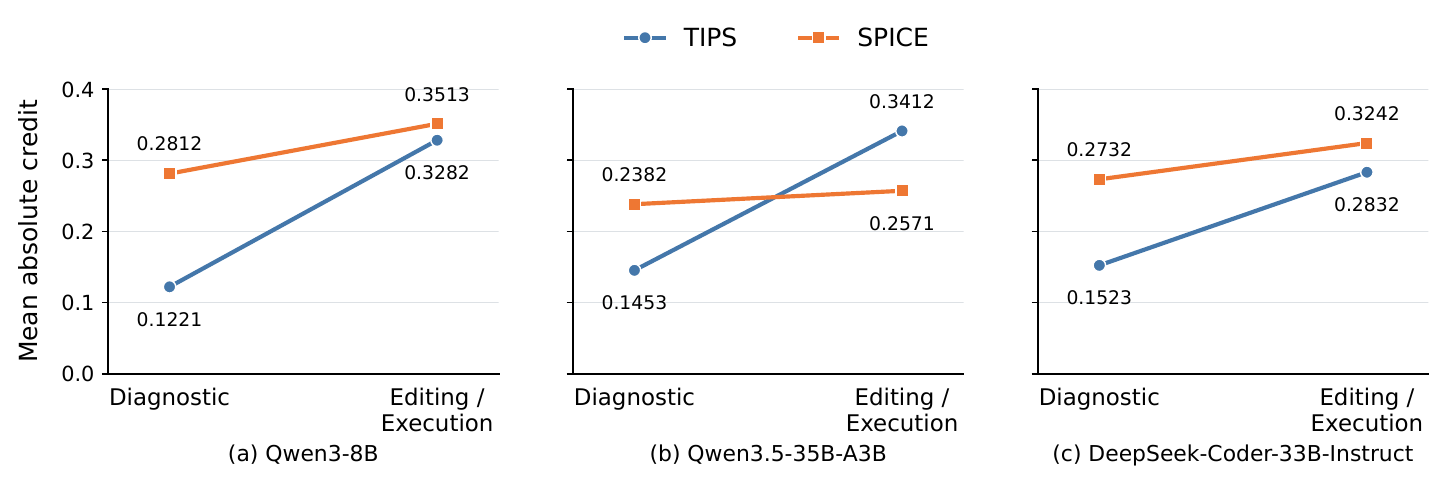}
\caption{\textbf{Feedback magnitude on fixed SFT trajectories.} Mean absolute unscaled scores: TIPS potential differences and \spice{} $d_{i,s}$. Panels use frozen, task-unadapted policy checkpoints; raw scales differ. Editing/execution includes code edits and execution. The plot compares feedback magnitudes; signed scores and diagnostic usefulness are assessed in Appendix~\ref{app:diagnostic_usefulness}.}
\label{fig:credit_by_action_type}
\end{figure}

\noindent\textbf{Feedback magnitude across action types.} Figure~\ref{fig:credit_by_action_type} reports mean absolute segment scores on fixed SFT trajectories, using three frozen policy checkpoints and a shared solution bank. The diagnostic-to-editing/execution ratio is 0.37--0.54 for TIPS versus 0.80--0.93 for \spice{}, indicating a smaller relative magnitude gap. Section~\ref{sec:ablations} compares success after training with each objective.

\noindent\textbf{Higher SPICE scores identify useful diagnostic calls.} SPICE and TIPS score the same calls using the same frozen 35B SFT checkpoint and reference summaries. SPICE ranks useful calls above redundant or irrelevant calls more often than TIPS does. Their respective AUROCs are \textbf{0.779 and 0.512 on ID}, and \textbf{0.713 and 0.496 on OOD} (Appendix~\ref{app:diagnostic_usefulness}). Under the SPICE policy, masking the information returned by high-scoring calls reduces success by \textbf{26.1 percentage points}, compared with \textbf{5.0} for low-scoring calls (Appendix~\ref{app:score_masking}).

\subsection{Ablation Studies}
\label{sec:ablations}

For Q4, we follow the pipeline from guidance and SFT to RL and reward design. Training-time tool dropout is reported in Appendix~\ref{app:supplementary_ablations}.

\begin{figure}[t]
\centering
\begin{minipage}[t]{0.49\linewidth}
\centering
\includegraphics[width=\linewidth]{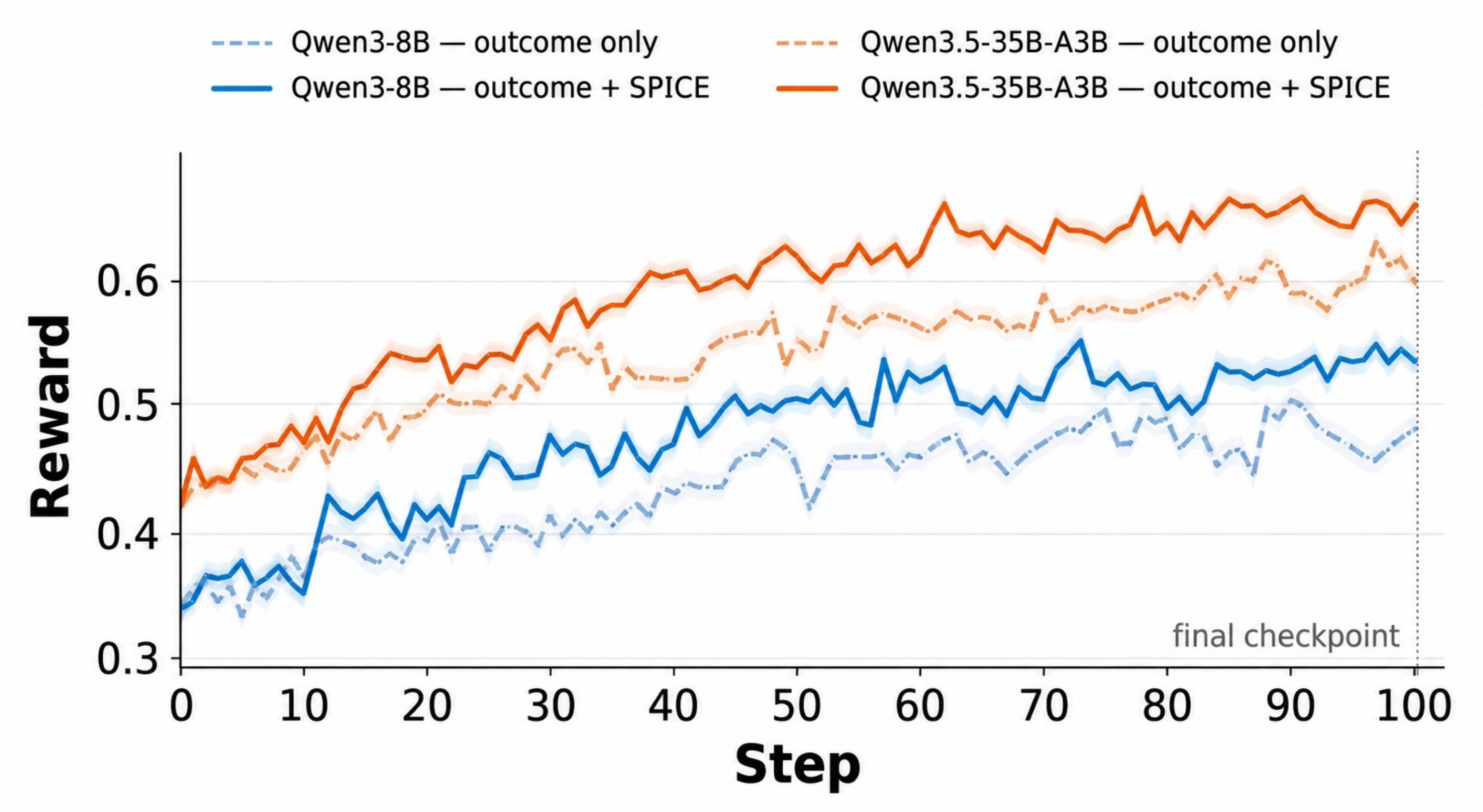}
\vspace{-2pt}\par\textbf{(a)} Training reward dynamics
\end{minipage}\hfill
\begin{minipage}[t]{0.49\linewidth}
\centering
\includegraphics[width=\linewidth]{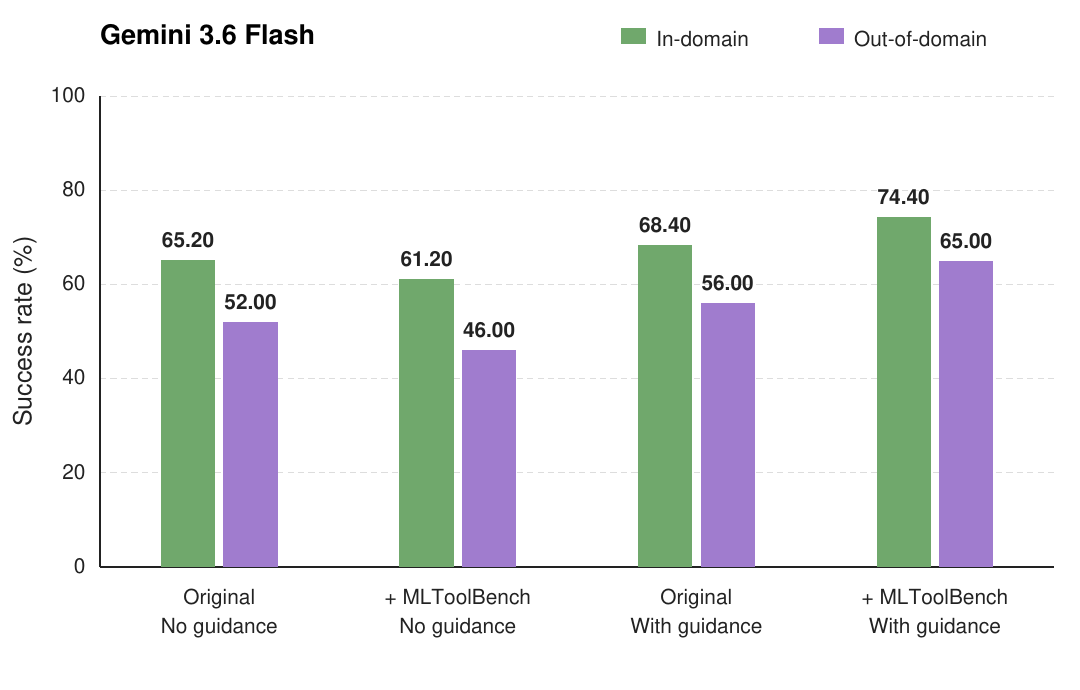}
\vspace{-2pt}\par\textbf{(b)} Guidance and diagnostic access
\end{minipage}
\caption{\textbf{Training and guidance ablations.} (a) Mean binary outcome reward $R_i$ for 8B (blue) and 35B (orange): outcome-only GRPO (dashed) and SPICE (solid). Both curves exclude shaping rewards. Final checkpoint: step 100. (b) Gemini ID/OOD success (\%); unguided results match Table~\ref{tab:main_results}. Training outcome success is distinct from held-out success.}
\label{fig:ablation_combined}
\label{fig:reward_dynamics}
\label{fig:learning_analysis}
\end{figure}

\noindent\textbf{Guidance and diagnostic access.} We first examine guidance for Gemini (Figure~\ref{fig:learning_analysis}). Diagnostics alone reduce ID success from 65.2\% to 61.2\%, whereas guidance alone raises it to 68.4\%. Combining both reaches 74.4\%. OOD success follows the same pattern: 52\%, 46\%, 56\%, and 65\%, respectively. This motivates teaching inspection strategies alongside tool access. Guided Gemini is a privileged reference receiving solution-informed hints without environment feedback (Appendix~\ref{app:agent_prompts}).

\noindent\textbf{Tool-use supervision.} We next examine SFT. With the augmented evaluation interface held fixed, tool-enabled demonstrations improve ID/OOD success over demonstrations collected without MLToolBench by 13.2/12 points for 8B and 5.2/9 for 35B (Appendix Figure~\ref{fig:sft_ablations}). This comparison evaluates the demonstration pipeline, including collection-time guidance, with equal raw trajectory budgets before filtering. Adding an unguided, tool-enabled collection condition for 35B separates two increments: tool-enabled collection without guidance adds 2.0/3.0 ID/OOD points, and guidance with tool access fixed adds a further 3.2/6.0 points (Appendix~\ref{app:collection_ablation}). 

\noindent\textbf{SPICE improves over matched RL objectives.} RL with \spice{} improves over full SFT by 14/4 ID/OOD points for 8B and 24.8/7 for 35B. Matched RL comparisons hold initialization, data, tools, and optimization settings fixed. SPICE exceeds outcome-only RL by 13.2/3 points for 8B and 21.2/4 for 35B, and TIPS by 8/1 and 14/3 points, respectively (Appendix Figure~\ref{fig:rl_ablations}). TIPS and SPICE share solution summaries, controlling reference content while changing how it supplies turn-level feedback. The 35B ID gaps correspond to 53 additional successes over outcome-only RL and 35 over TIPS among 250 evaluations. Across three independent 35B RL seeds, SPICE averages 68.7\%/48.0\% ID/OOD success versus 47.9\%/43.0\% for outcome-only RL (Appendix~\ref{app:rl_seeds}).

\noindent\textbf{Assigning each action its own score improves success.} We shuffle SPICE scores among actions from the same task and tool category, preserving the distribution of signed scores in each group. The 35B policy trained with shuffled scores reaches 59.2\% ID and 45\% OOD success. Using each action's own score raises success to 69.2\% and 48\%, gains of 10 and 3 points (Appendix Table~\ref{tab:summary_completeness}). Even shuffled scores improve on outcome-only RL by 11.2 points on ID and 1 point on OOD. A separate test changes the reference content while keeping the scoring rule fixed. With reduced solution summaries, success falls to 67.2\% and 41\%.

\noindent\textbf{$\beta$ sensitivity.} Pilot calibration of $\beta$ to small, medium, and large shaping-reward ranges yields 67.2\%, 69.2\%, and 65.6\% ID success for 35B, respectively, favoring medium shaping (Appendix~\ref{app:reward_magnitude}).

\noindent\textbf{Additional controls.} We compare reusable guides with training on the original 35B checkpoint and vary the information supplied by the guides (Appendix~\ref{app:reusable_guide_results}). We also analyze failures after adding diagnostics and test RL without the SFT cold start (Appendices~\ref{app:unadapted_failures} and~\ref{app:direct_rl}). Under the outcome-only training-time budget, including initial-summary preparation, SPICE reaches 65.2\%/46.0\% ID/OOD success, compared with 48.0\%/44.0\% for outcome-only RL (Appendix~\ref{app:runtime}).

\section{Related Work}

\noindent\textbf{ML Development Agents.} MLAgentBench, MLE-bench, MLGym, and DeltaML-Bench provide executable evaluation \citep{huang2024mlagentbench,chan2025mlebench,nathani2025mlgym,moukpe2026deltaml}. AIDE and AIRA-dojo organize code search, while ARTS and Matryoshka structure diagnosis and exploration \citep{jiang2025aide,toledo2025aira,juneja2026arts,qiang2026matryoshka}. Related efforts develop interaction-trained agents \citep{liu2025mlagent,yang2026rlmle} and scalable tasks and demonstrations in MLE-Dojo, MLE-Smith, SandMLE, and ML-AutoResearch \citep{qiang2025mledojo,qiang2025mlesmith,zhou2026sandmle,cai2026mlautoresearch}. We complement these efforts with learnable diagnostic actions and privileged-context feedback within an existing scaffold.

\noindent\textbf{Tool Use Learning.} Toolformer, API-Bank, and ToolLLM study tool-selection and call-generation supervision \citep{schick2023toolformer,li2023apibank,qin2023toolllm}; ToolSandbox and $\tau$-bench evaluate dependencies across calls \citep{lu2024toolsandbox,yao2024taubench}. ReTool and Tool-Star learn interaction strategies with outcome and hierarchical rewards \citep{feng2025retool,dong2025toolstar}. We separately evaluate diagnostic access and tool-enabled supervision for evidence acquisition and use.

\noindent\textbf{Intermediate Feedback and Credit Assignment.} TIPS tracks correct-answer likelihood changes, and TACO uses before/after answer probes \citep{xie2026tips,feng2026taco}. Agent0-VL combines verifier feedback with self-repair \citep{liu2026agent0vl}; other methods use semantic role rewards or hindsight values \citep{xu2026triage,tan2026hcapo}. Self-Distilled RLVR uses positive likelihood-ratio weights without reversing advantage signs, while SEED uses hindsight skills for sampled-token learning \citep{yang2026rlsd,wu2026seed}. SPICE instead adds signed action-support scores as turn rewards, retaining an executable outcome evaluator without a learned verifier (Appendix~\ref{app:dynamic_credit}).

\section{Conclusion}

We built \toolbench{} from failures we observed in ML development tasks, then trained two Qwen models to choose diagnostic tools and act on their findings. Both improved on ID and OOD tasks. Starting from the same SFT checkpoint, training with SPICE achieved higher success than training with TIPS or outcome-only RL. The advantage over outcome-only RL persisted across three 35B training seeds, and SPICE also scored highest among our checkpoints on MLS-Bench-Lite. Two findings help explain what SPICE rewards. High-scoring calls were more often judged useful, and withholding their results reduced final task success more than withholding the results of low-scoring calls.

\section*{Limitations}

We test transfer to external tasks on 30 MLS-Bench-Lite instances and transfer to a third-party diagnostic interface on five of our tabular tasks. All three RL seeds start from the same SFT checkpoint, so they measure variability during RL. Two LLM judges labeled diagnostic usefulness, and human annotators checked 20 trajectories. The masking experiment measures how a fixed policy responds when information is withheld at a particular call. The high- and low-score groups can differ in both task state and tool choice. Throughout training, SPICE supplements the terminal outcome reward (Appendix~\ref{app:dynamic_credit}).

\section*{Reproducibility Statement}
Section~\ref{sec:experimental_setup} specifies the evaluation splits and success criterion. Appendix~\ref{app:experimental_protocol} summarizes task budgets, the success-rate metric, model checkpoints, and RL settings; the remaining appendices provide ablations, tools, and SFT details. The terminal success criterion, repeated-evaluation protocol, and outcome-only control are shared across the corresponding comparisons.

\section*{AI Use Statement}
Generative AI tools assisted literature search and summarization, exploration and refinement of research ideas and task designs, experimental design and interpretation of results, manuscript drafting and language editing, figure creation, and code development and debugging. Gemini 3.6 Flash was also used to generate training demonstrations, as described in Section~\ref{sec:guided_sft}. AI-assisted suggestions do not replace empirical validation. The authors retain responsibility for the final text, citations, methodological claims, code, figures, and reported results.

\bibliography{references}
\bibliographystyle{references}

\clearpage
\beginappendix

\section{Experimental Protocol and Reproducibility}
\label{app:experimental_protocol}

\subsection{Task Splits and Evaluation}
\label{app:task_collection}
\label{app:evaluation_metrics}
SFT and RL use the same 80 training tasks; all 35 evaluation tasks are held out from demonstration collection and training solution banks. Table~\ref{tab:setup_counts} gives the collection and evaluation budgets. Each rollout ends after 50 agent steps or one hour, whichever occurs first. Original (standard MLAgentBench tools only) and + MLToolBench (the same tools plus diagnostics) share the same tasks and execution budgets. Each condition uses ten evaluation rollouts per task with the same fixed model or checkpoint. These are repeated task executions, not independent RL training runs.

\begin{table}[!htbp]
\centering\small
\begin{tabularx}{\linewidth}{@{}lrrY@{}}
\toprule
Partition & Instances & Runs per task & Total / purpose \\
\midrule
Training & 80 & 100 & 8,000 raw demonstrations per SFT condition \\
ID evaluation & 25 & 10 & 250 evaluation runs per condition \\
OOD evaluation & 10 & 10 & 100 evaluation runs per condition \\
\bottomrule
\end{tabularx}
\caption{Collection and evaluation budgets. RL collects fresh on-policy rollouts.}
\label{tab:setup_counts}
\end{table}

Our success criterion is adapted from, but not identical to, MLAgentBench's relative-improvement criterion \citep{huang2024mlagentbench}. Direct division by the baseline is undefined at zero and reverses the improvement direction for negative baselines. We instead compare the direction-adjusted absolute gain against a threshold based on the baseline magnitude, with a minimum absolute-gain floor to avoid counting negligible changes near zero. Let $m_{x,r}$ and $b_x$ be the final and starter metrics and $\sigma_x\in\{+1,-1\}$ indicate the favorable direction. Define
\begin{equation}
U_{x,r}=\mathbf 1\!\left[V_{x,r}=1\land m_{x,r}\text{ finite}\land
\sigma_x(m_{x,r}-b_x)>\max(0.10|b_x|,\delta)\right].
\label{eq:extended_success}
\end{equation}
We use $\delta=0.01$ as the minimum absolute gain in the metric values returned by the task evaluator. The floor is defined on each evaluator's fixed native metric scale and is not scale-invariant. We checked the starter baselines and evaluator-defined metric bounds for all retained synthetic tasks; none has a success threshold rendered unattainable by these bounds. Valid zero and negative scores are retained. Non-finite final scores fail; a non-finite baseline requires repairing the evaluator configuration before evaluation. For $b_x>0$ and $0.10b_x\geq\delta$, this reduces to the original strict 10\% relative-gain rule. For $0<b_x<0.1$, the absolute floor makes our criterion stricter than a 10\% relative-gain threshold. We therefore report results under this extended rule rather than claim exact equivalence to MLAgentBench preprocessing. We report
\begin{equation}
\mathrm{SR}(\mathcal T)=\frac{100\%}{10|\mathcal T|}\sum_{x\in\mathcal T}\sum_{r=1}^{10}U_{x,r}.
\label{eq:success_rate}
\end{equation}
Each task has ten runs and equal weight. The observation intervention also includes all 250 ID rollouts per policy. Training-task test scripts compare submissions and starter baselines on the same partition and supply binary terminal rewards using the same criterion. SFT demonstration filtering is separate. Outcome rewards, shaped returns, and five-rollout tool coverage are separate measures.

\subsection{Model Checkpoints}
\label{app:training_config}
\label{app:model_identifiers}
The reported API model names are \nolinkurl{gemini-3.6-flash} and \nolinkurl{claude-sonnet-5}. Open checkpoints are \nolinkurl{Qwen/Qwen3-8B}, \nolinkurl{Qwen/Qwen3.5-35B-A3B}, and \nolinkurl{deepseek-ai/deepseek-coder-33b-instruct}. Qwen3-8B is the official post-trained model; ``unadapted'' means before our task-specific SFT and RL. The coding baseline uses the pretrained \href{https://huggingface.co/Qwen/Qwen2.5-Coder-14B}{\nolinkurl{Qwen/Qwen2.5-Coder-14B}} checkpoint.

\subsection{RL Configuration}
\label{app:implementation_status}
RL starts from each backbone's tool-enabled full-SFT checkpoint. Table~\ref{tab:rl_repro_config} lists its settings; reported results use the final checkpoint after 100 optimizer updates. SFT collection and optimization appear in Appendix~\ref{app:sft_details}, prompts in Appendix~\ref{app:agent_prompts}, and scoring details in Appendix~\ref{app:dynamic_credit}.

\begin{table}[H]
\centering\small
\begin{tabularx}{\linewidth}{@{}L{5.0cm}Y@{}}
\toprule
RL configuration & Value \\
\midrule
Optimizer updates / reported checkpoint & 100 / final step 100 \\
Rollout batch & 8 tasks $\times$ 8 rollouts = 64 trajectories \\
Optimizer / learning rate / scheduler & AdamW / $1\times10^{-6}$ / cosine \\
Minibatch size / update epochs / accumulation & 128 / 1 / 1 \\
Advantage estimation & Task-group mean centering; no standard-deviation scaling \\
Clip parameter / KL coefficient & 0.2 / 0 (KL disabled) \\
Terminal reward $R_i$ & Binary success under Equation~\ref{eq:extended_success} \\
Shaping calibration & Pilot mean absolute turn reward targeted to $[0.05,0.3]$ (Appendix~\ref{app:shaping_calibration}) \\
Updated policy model & Same frozen rollout-policy snapshot in both branches \\
Privileged solutions & Up to 8 verified references per task; new solutions enter the next update \\
Training-time tool dropout (ablation) & 20\%, sampled once at rollout start and fixed thereafter \\
Sampling temperature / top-$p$ / token limit & 1.0 / 1.0 / 1024 \\
Hardware / precision & 8$\times$ H200 / BF16 \\
\bottomrule
\end{tabularx}
\caption{RL configuration and shaping-scale calibration protocol. Rollout collection and optimization batch sizes are distinct. The calibration range refers to reward magnitude, not the coefficient.}
\label{tab:rl_repro_config}
\end{table}

\section{Ablation and Behavioral Analysis}
\subsection{Additional Ablation Results}
\label{app:supplementary_ablations}

Success rates use Equation~\ref{eq:success_rate}, with 250 ID and 100 OOD runs per condition; the observation intervention reports ID only. Conditions shared with Table~\ref{tab:main_results} use the same results.

\begin{figure}[H]
\centering
\includegraphics[width=\linewidth]{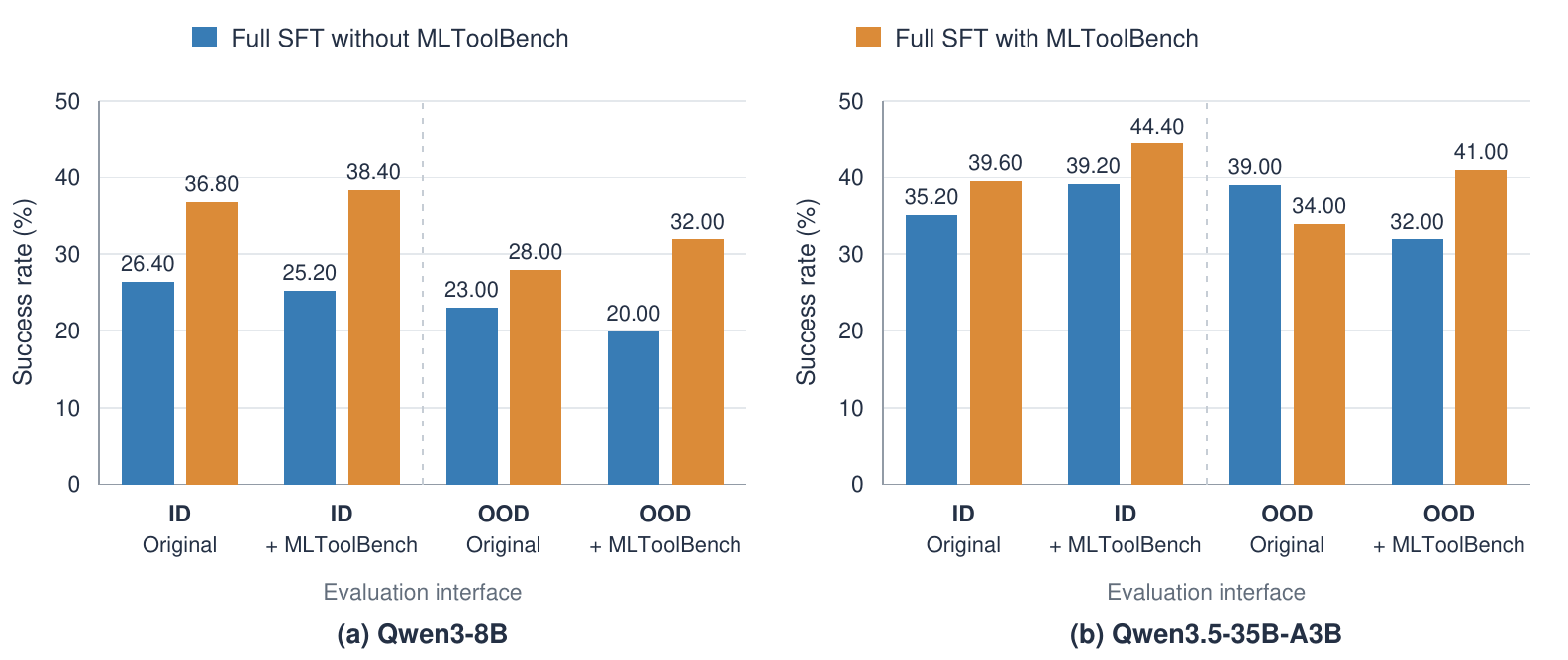}
\caption{\textbf{Full-SFT training conditions and evaluation interfaces.} Each group compares SFT with and without MLToolBench demonstrations at a fixed evaluation interface. Original uses standard MLAgentBench tools without MLToolBench diagnostics; + MLToolBench adds these diagnostics at evaluation. Both panels share a scale and reproduce the tool-enabled SFT results in Table~\ref{tab:main_results}.}
\label{fig:sft_ablations}
\end{figure}

\noindent\textbf{SFT ablations.} Figure~\ref{fig:sft_ablations}(a,b) compares full SFT using demonstrations collected with or without MLToolBench, each starting from 8,000 raw trajectories before filtering and deduplication. Both policies are evaluated on Original and + MLToolBench, as defined in Section~\ref{sec:experimental_setup}. The no-MLToolBench policy on Original provides a matched-interface baseline; the tool-trained policy on Original measures diagnostic removal. Tool-enabled SFT results reproduce Table~\ref{tab:main_results}.

\begin{figure}[H]
\centering
\includegraphics[width=\linewidth]{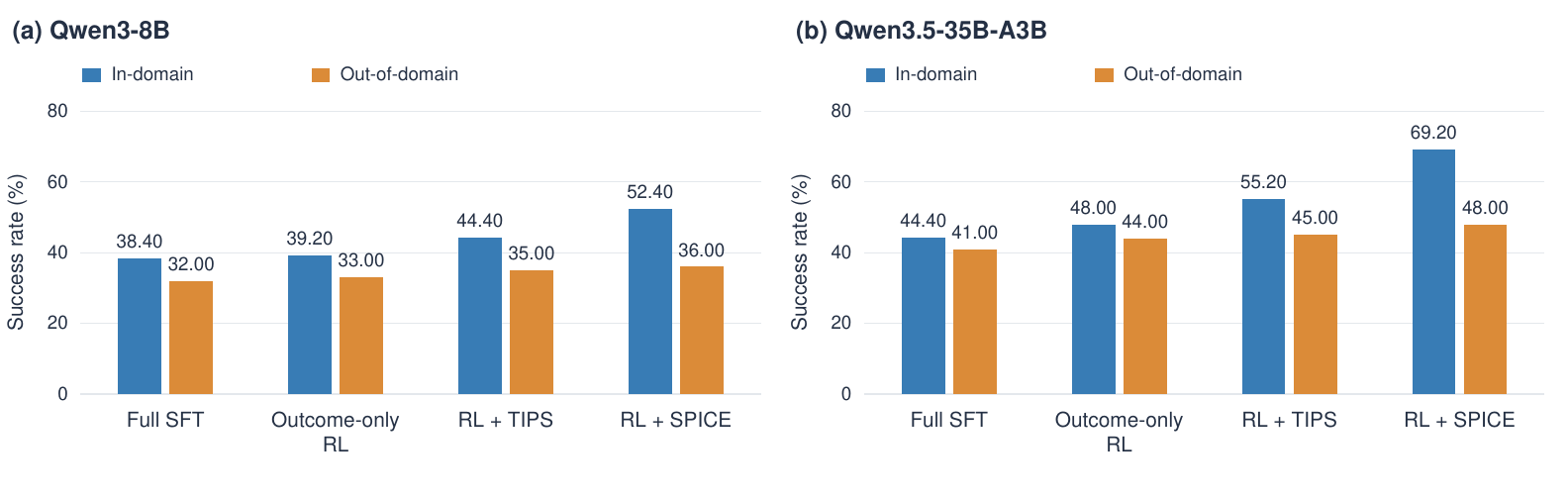}
\caption{\textbf{RL-stage and reward comparisons with MLToolBench.} Full SFT and RL + \spice{} values reproduce Table~\ref{tab:main_results}. Outcome-only RL ($\beta=0$), RL + TIPS, and RL + \spice{} are shown for both backbones. Gains over SFT measure the entire RL stage.}
\label{fig:rl_ablations}
\end{figure}

\noindent\textbf{RL and turn-level shaping.} Figure~\ref{fig:rl_ablations} compares full SFT with its RL continuation under \toolbench{}; original-interface results remain in Table~\ref{tab:main_results}. The 8B outcome-only control ($\beta=0$) reports 39.20\% ID and 33.00\% OOD success, versus 52.40\% and 36.00\% with \spice{}. For each backbone, the three RL conditions share the SFT initialization, training data, tool interface, rollout budget, optimization settings, and binary outcome reward. Outcome-only disables shaping with $\beta=0$, while TIPS and \spice{} use their respective shaping methods. The 35B outcome-only control reports 48.00\% ID and 44.00\% OOD success, versus 69.20\% and 48.00\% with \spice{}. RL with TIPS shaping achieves 44.40\% ID (111/250) and 35.00\% OOD (35/100) for 8B, and 55.20\% ID (138/250) and 45.00\% OOD (45/100) for 35B.

\noindent\textbf{Training-time tool dropout.} Dropping 20\% of diagnostics once at each training rollout\textquotesingle s start, with availability fixed thereafter, changes ID/OOD success from 52.4\%/36\% to 50\%/31\% for 8B and from 69.2\%/48\% to 58\%/36\% for 35B. Evaluation uses all tools: this tests incomplete training-time access, not robustness to missing tools at inference.

\label{app:reward_magnitude}
\noindent\textbf{$\beta$ sensitivity.} For Qwen3.5-35B-A3B with \toolbench{}, we vary $\beta$ to calibrate pilot mean absolute turn-level shaping rewards to small ($[0.001,0.05]$), medium ($[0.05,0.3]$), and large ($[0.3,1.0]$) ranges, obtaining ID success rates of 67.2\%, 69.2\%, and 65.6\%, respectively. The medium range performs best, exceeding the alternatives by 2.0 and 3.6 percentage points. These intervals specify reward magnitudes, not $\beta$ values.

\noindent\textbf{Solution-summary completeness.} For 35B SPICE, the reduced-summary condition yields 168/250 ID successes and 41/100 OOD successes, compared with 173/250 and 48/100 using the full candidate summaries (Table~\ref{tab:summary_completeness}). The reductions are 2.0 and 7.0 percentage points. Partial summaries retain an ID gain over outcome-only RL (67.2\% versus 48.0\%), but OOD success is lower (41\% versus 44\%).

\noindent\textbf{Action--feedback correspondence.} We shuffle signed SPICE scores among actions within the same task and tool category (Diagnostic or Editing / Execution), preserving each permutation group's score distribution while disrupting its action correspondence. The matched RL control reaches 59.2\% ID (148/250) and 45.0\% OOD (45/100), versus 69.2\%/48.0\% for SPICE. The shuffled control retains an 11.2/1.0-point gain over outcome-only RL. The additional 10.0/3.0-point gain from preserving action--feedback correspondence supports the value of targeted feedback.

\begin{table}[H]
\centering\small
\begin{tabularx}{\linewidth}{@{}Ycc@{}}
\toprule
35B training condition & ID success (\%) & OOD success (\%) \\
\midrule
Outcome-only RL & 48.00 (120/250) & 44.00 (44/100) \\
SPICE, full solution summaries & 69.20 (173/250) & 48.00 (48/100) \\
SPICE, reduced solution summaries & 67.20 (168/250) & 41.00 (41/100) \\
SPICE, shuffled scores & 59.20 (148/250) & 45.00 (45/100) \\
\bottomrule
\end{tabularx}
\caption{\textbf{Privileged-summary and score-assignment ablations.} Success rates for Qwen3.5-35B-A3B with MLToolBench. Shuffling permutes scores within tasks and tool categories. Full-summary SPICE reproduces the main result.}
\label{tab:summary_completeness}
\end{table}

\subsection{Behavioral Analysis Details}
\label{app:tool_behavior}

\begin{figure}[H]
\centering
\includegraphics[width=\linewidth]{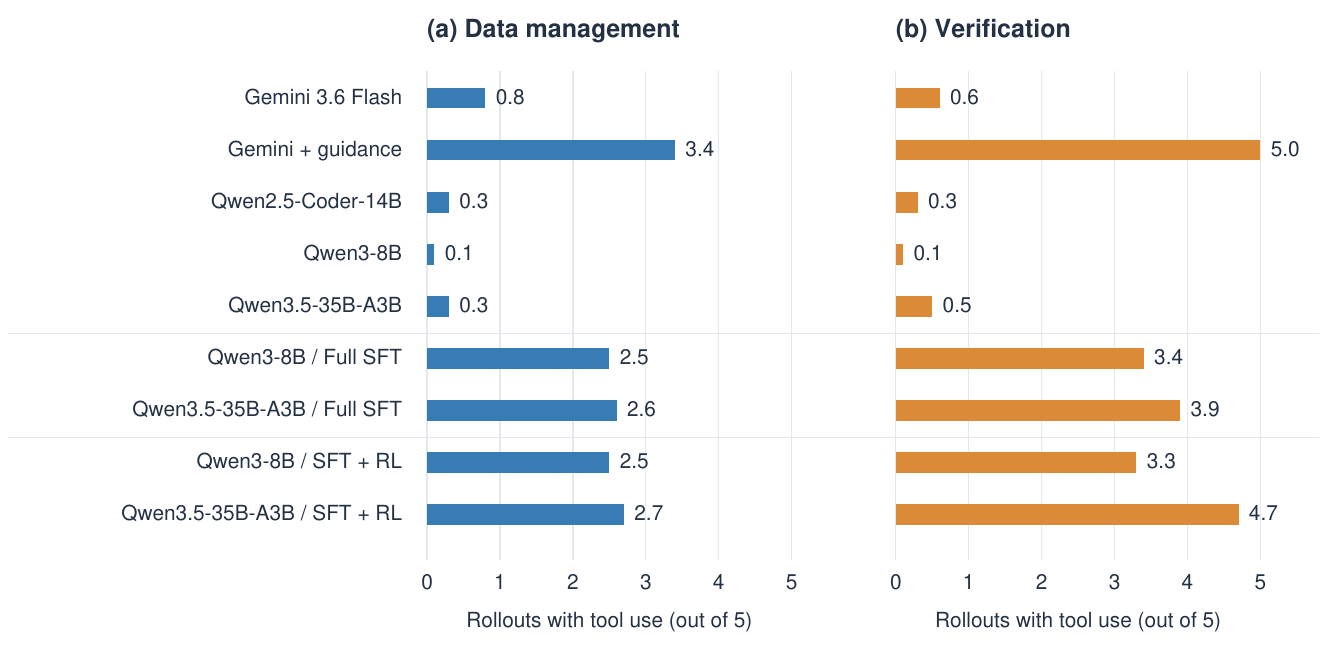}
\caption{\textbf{Tool-use coverage by model and training condition.} Mean number of rollouts, out of five per task, invoking data-management or verification tools at least once. Separators distinguish unadapted baselines, full SFT, and full SFT + RL with \spice{}.}
\label{fig:tool_coverage}
\end{figure}

Figure~\ref{fig:tool_coverage} reports tool-use coverage. For behavior-analysis tasks $\mathcal B$ and category $d$'s tool set $\mathcal U_d$, we compute
\begin{equation}
\mathrm{Coverage}_d=\frac{1}{|\mathcal B|}\sum_{x\in\mathcal B}
\sum_{r=1}^{5}\mathbf 1\!\left[\exists s:\ \operatorname{tool}(a_{x,r,s})\in\mathcal U_d\right].
\label{eq:tool_coverage}
\end{equation}
Each of five complete rollouts contributes at most one count per category, regardless of repeated calls. Task-averaged values range from zero to five; 4.7 corresponds to 94\% invocation coverage. This measures invocation; task success is evaluated separately with ten rollouts per task.

\noindent\textbf{Call frequency and tool diversity.} Table~\ref{tab:tool_use_statistics} counts MLToolBench calls and distinct tool IDs per rollout. It excludes the scaffold's standard editing and execution actions. The 35B SPICE policy uses eight distinct tools in 11 calls on average, compared with two in nine calls for outcome-only RL and three in ten calls for TIPS. Appendix~\ref{app:diagnostic_usefulness} evaluates whether individual calls, including repeated checks, provide useful information.

\begin{table}[H]
\centering\small
\begin{tabularx}{\linewidth}{@{}Ycccc@{}}
\toprule
& \multicolumn{2}{c}{Coverage (out of 5)} & \multicolumn{2}{c}{Mean per rollout} \\
\cmidrule(lr){2-3}\cmidrule(lr){4-5}
Model / training & Data management & Verification & Calls & Distinct tools \\
\midrule
8B Full SFT & 2.5 & 3.4 & 5 & 2 \\
8B RL + \spice{} & 2.5 & 3.3 & 11 & 6 \\
35B Full SFT & 2.6 & 3.9 & 7 & 4 \\
35B Outcome-only RL & 0.8 & 1.3 & 9 & 2 \\
35B RL + TIPS & 1.4 & 1.6 & 10 & 3 \\
35B RL + \spice{} & 2.7 & 4.7 & 11 & 8 \\
\bottomrule
\end{tabularx}
\caption{\textbf{MLToolBench invocation, frequency, and diversity.} Coverage counts rollouts invoking each category at least once; calls count invocations, and distinct tools count unique MLToolBench tool IDs within a rollout. Editing/execution actions are excluded.}
\label{tab:tool_use_statistics}
\end{table}

\subsection{Paired Diagnostic-Observation Intervention}
\label{app:observation_intervention}
\noindent\textbf{Evaluation setup.} We compare matched 35B outcome-only and \spice{} policies on all 250 ID rollouts per policy (ten per task). Each policy generates its own trajectories. True is its unmodified evaluation, matching the corresponding main or RL-ablation result. Rollouts without diagnostic calls are also included in this evaluation.

\noindent\textbf{Intervention.} For a trajectory with MLToolBench diagnostic calls, we randomly select one read-only diagnostic return and branch from its saved history and workspace. True receives the original return. Masked preserves its response structure and execution status but marks measurements, conclusions, and derived suggestions unavailable. Paraphrased retains facts and values while changing presentation. Branches use paired seeds, identical remaining budgets and decoding settings; subsequent tools operate normally. Trajectories with no diagnostic call remain unchanged in all three arms and contribute their original outcomes. The task test script evaluates final success.

\noindent\textbf{Interpretation.} All three arms use the same 250-rollout denominator within each policy. Differences measure the effect of the observation intervention across the full ID evaluation, including unchanged no-call trajectories. Appendix~\ref{app:score_masking} separately examines the relationship between SPICE scores and the effect of masking.

\begin{table}[H]
\centering\small
\begin{tabularx}{\linewidth}{@{}lYc@{}}
\toprule
35B training objective & Observation & ID success (\%) \\
\midrule
\multirow{3}{*}{Outcome-only RL} & True & 48.00 \\
 & Masked & 46.40 \\
 & Paraphrased & 47.60 \\
\midrule
\multirow{3}{*}{RL + \spice{}} & True & 69.20 \\
 & Masked & 54.40 \\
 & Paraphrased & 68.40 \\
\bottomrule
\end{tabularx}
\caption{\textbf{Diagnostic-observation intervention on ID tasks.} Both 35B policies retain all tools. All 250 ID rollouts are included. We randomly intervene on one diagnostic return where available; no-call trajectories retain their original outcomes in all arms.}
\label{tab:observation_intervention}
\end{table}

\section{\toolbench{} Interface Details}
\label{app:toolbench_details}

\toolbench{} contains \textbf{67 registered diagnostic tools}: seven shared tools plus 12 tabular, 15 time-series, 17 NLP, eight vision, and eight graph tools. These additions are separate from the scaffold's 17 base actions. Tables~\ref{tab:appendix_general_tools} and~\ref{tab:appendix_domain_tools} give the complete inventory. The three functional groups in the main text describe purposes, whereas the five domain modules organize implementations; time-series tools also diagnose training curves from other domains.

\noindent\textbf{Interface and exposure.} Calls specify workspace files, logs, data fields, or model references and return measurements, flags, or heuristic suggestions. Registered tools are not all listed in every prompt: explicit tool selections or task-to-domain mappings determine prompt exposure, with additional diagnostics omitted by default. Prompt exposure does not itself enforce execution permissions.

\noindent\textbf{Execution and interpretation.} Most tools inspect existing data or artifacts. \emph{Verify Hyperparam Effective} performs static analysis rather than verifying runtime values; \emph{Trace Tensor Shapes} executes a dummy forward pass; \emph{Estimate Peak Memory} briefly launches training and can inherit the script's side effects. Suggestions are not applied automatically. Heuristic scores are diagnostic proxies, not correctness guarantees, and model-dependent probes require the specified model or tokenizer. The tool registry does not itself enforce held-out-label isolation.

\begingroup
\footnotesize
\setlength{\tabcolsep}{5pt}
\renewcommand{\arraystretch}{1.12}
\begin{longtable}{@{}L{4.7cm}L{8.5cm}@{}}
\toprule Tool & Returned evidence / behavior \\
\midrule
\endfirsthead
\multicolumn{2}{l}{\small Table \thetable{} (continued)}\\
\toprule Tool & Returned evidence / behavior \\
\midrule
\endhead
\midrule
\multicolumn{2}{r}{\footnotesize Continued on next page}\\
\endfoot
\bottomrule
\caption{Seven shared diagnostic tools. Names match the registered action interface.}\label{tab:appendix_general_tools}\\
\endlastfoot
\emph{Discover Log Metrics} & Metric names, sample values, and occurrence counts in logs or captured stdout. \\
\emph{Extract Training Metric Series} & Value--step series for requested logged metrics. \\
\emph{Verify Hyperparam Effective} & Static parameter-definition sites, conflicting values, and environment overrides. \\
\emph{Estimate Peak Memory} & Observed peak GPU memory during a short training-script probe; OOM status. \\
\emph{Trace Tensor Shapes} & Per-layer tensor shapes or failure traces from a CPU dummy forward pass. \\
\emph{Check Config Consistency} & Rule-based checks of configuration relations and incompatible settings. \\
\emph{Validate Output Format} & Record-format compliance, missing fields, and length or pattern violations. \\
\end{longtable}
\endgroup

\begingroup
\footnotesize
\setlength{\tabcolsep}{5pt}
\renewcommand{\arraystretch}{1.12}
\begin{longtable}{@{}L{4.7cm}L{8.5cm}@{}}
\toprule Tool & Returned evidence / behavior \\
\midrule
\endfirsthead
\multicolumn{2}{l}{\small Table \thetable{} (continued)}\\
\toprule Tool & Returned evidence / behavior \\
\midrule
\endhead
\midrule
\multicolumn{2}{r}{\footnotesize Continued on next page}\\
\endfoot
\bottomrule
\caption{Complete domain-module inventory: 60 tools across five modules. Descriptions summarize implemented checks, including heuristic suggestions.}\label{tab:appendix_domain_tools}\\
\endlastfoot
\multicolumn{2}{@{}l}{\textbf{Tabular (12 tools)}}\\*
\emph{Detect Missing Values} & Per-column missingness and suggested imputation strategies. \\
\emph{Identify Feature Types} & Inferred column types, unique counts, and ID/high-cardinality flags. \\
\emph{Detect Label Imbalance} & Class counts, imbalance ratios, and resampling suggestions. \\
\emph{Check Feature Normalization} & Numeric ranges, means, standard deviations, and scaling suggestions. \\
\emph{Detect Outliers} & Per-column outlier counts and IQR bounds. \\
\emph{Correlation Analysis} & Highly correlated feature pairs indicating possible redundancy. \\
\emph{Cardinality Check} & Categorical unique counts and suggested encodings. \\
\emph{Feature Skewness Checker} & Numeric-feature skewness and suggested transformations. \\
\emph{Duplicate Row Detector} & Exact duplicate-row counts and proportions. \\
\emph{Detect Train/Test Overlap} & Row overlap between two supplied datasets on selected columns. \\
\emph{Encoding Strategy Advisor} & Encoding recommendations based on type, cardinality, and sample size. \\
\emph{Feature Interaction Detector} & Numeric feature-product pairs ranked by estimated mutual-information gain. \\
\addlinespace[4pt]
\multicolumn{2}{@{}l}{\textbf{Time series (15 tools)}}\\*
\emph{Z-Score Anomaly Detector} & Global standardized outlier positions and magnitudes. \\
\emph{IQR Anomaly Detector} & Outlier positions and interquartile-range bounds. \\
\emph{Moving Window Anomaly Detector} & Locally unusual points relative to a moving window. \\
\emph{Trend Detection} & Linear slope, fit statistics, and rank-based trend evidence. \\
\emph{Plateau Detector} & Tail slope and relative change indicating a plateau. \\
\emph{Convergence Estimator} & Early/late volatility and heuristic convergence indicators. \\
\emph{Change Point Detection} & Largest detected mean-shift location and before/after means. \\
\emph{Spike Detector} & Abrupt step-to-step changes relative to typical differences. \\
\emph{Volatility Analysis} & Rolling and overall variability summaries. \\
\emph{NaN/Inf Health Detector} & Non-finite values and large-magnitude warnings. \\
\emph{Seasonality Detector} & Dominant frequency-derived period and periodicity strength. \\
\emph{Autocorrelation Function (ACF)} & Autocorrelation by lag and significant-lag summaries. \\
\emph{Partial Autocorrelation Function (PACF)} & Partial autocorrelation and suggested autoregressive order. \\
\emph{Stationarity Test (ADF)} & ADF statistic, p-value, and stationarity/differencing suggestions. \\
\emph{Granger Causality Test} & Lagged predictive-dependence tests between two supplied series. \\
\addlinespace[4pt]
\multicolumn{2}{@{}l}{\textbf{NLP (17 tools)}}\\*
\emph{Repetition Detector} & Exact and n-gram-based near-duplicate text ratios. \\
\emph{Toxicity / Noise Filter} & Blank, markup, symbol-noise, and other heuristic text-quality flags. \\
\emph{Language Consistency Checker} & Detected language proportions and mixed-language indicators. \\
\emph{Instruction-Response Coherence} & Lexical-overlap proxy for instruction--response agreement. \\
\emph{Token Length Distribution} & Tokenizer-based length statistics and potential truncation rates. \\
\emph{Vocabulary Coverage Checker} & Unknown-token proportion for a supplied tokenizer. \\
\emph{Prompt Template Validator} & Presence of expected prompt markers and compliance rates. \\
\emph{Response Quality Scorer} & Heuristic length and repetition indicators for responses. \\
\emph{Topic Distribution Analyzer} & TF-IDF clustering summaries and topic concentration. \\
\emph{Difficulty Distribution Estimator} & Length/complexity-based difficulty proxies and their distribution. \\
\emph{Data Source Mixing Analyzer} & Source counts, mixing ratios, and imbalance suggestions. \\
\emph{Perplexity-based Outlier Detector} & Reference-LM perplexity statistics and high/low outliers. \\
\emph{Reward Distribution Analyzer} & Reward-distribution statistics and discrimination indicators. \\
\emph{KL Reference Divergence Estimator} & Estimated policy--reference KL on supplied text and penalty suggestions. \\
\emph{Generation Diversity Checker} & Distinct n-grams, repeated outputs, and collapse indicators. \\
\emph{Safety Content Detector} & Keyword-based sensitive-content flags and matched terms. \\
\emph{Refusal Pattern Analyzer} & Refusal-phrase frequency and heuristic over-refusal indicators. \\
\addlinespace[4pt]
\multicolumn{2}{@{}l}{\textbf{Vision (8 tools)}}\\*
\emph{Class Balance Check} & Image-class counts and imbalance ratios. \\
\emph{Image Normalization Stats} & Pixel ranges, channel means/standard deviations, and normalization suggestions. \\
\emph{Image Shape Consistency} & Image-shape/channel distributions and mismatches. \\
\emph{Duplicate Image Detector} & Exact and near-duplicate image counts using image hashes. \\
\emph{Corrupt Image Detector} & Unreadable or corrupt image-file counts. \\
\emph{Augmentation Auditor} & Static training-script scan for transforms and suggested additions. \\
\emph{Train/Test Distribution Shift} & Differences in channel statistics between two supplied image sets. \\
\emph{Per-Class Accuracy Breakdown} & Per-class accuracy and confusions from predictions and supplied labels. \\
\addlinespace[4pt]
\multicolumn{2}{@{}l}{\textbf{Graph (8 tools)}}\\*
\emph{Graph Size Summary} & Node/edge counts, density, feature dimensions, and split sizes. \\
\emph{Degree Distribution} & Degree statistics, isolated nodes, and high-degree hubs. \\
\emph{Node Label Balance} & Class counts on training nodes when available, otherwise supplied labels. \\
\emph{Node Feature Health} & Feature scale, non-finite values, zero rows, and constant columns. \\
\emph{Connectivity Components} & Connected-component counts and largest-component fraction. \\
\emph{Self-loop \& Duplicate Edge Detector} & Self-loop/duplicate-edge counts and reverse-edge coverage. \\
\emph{Homophily Estimator} & Same-label edge fraction using supplied node labels. \\
\emph{Split Integrity Check} & Split sizes, index overlaps, and supplied-label class coverage. \\
\end{longtable}
\endgroup

\section{Training Details}
\label{app:training_details}
\subsection{Supervised Fine-Tuning}
\label{app:sft_details}

\noindent\textbf{Guidance construction and use.}
\label{app:teacher_prompt}
The guidance-construction LLM receives the training task, its reference solution, and the synthetic environment generation mechanism. For each instance, it connects the prediction target, data modality, starter configuration, and generation conditions to concrete diagnostic checks. Guidance specifies the evidence to collect, the interpretation of alternative outcomes, and the corresponding edits and validation checks; different instances of the same source task can therefore receive different guidance. The resulting guidance is appended to the demonstration model's task prompt and specifies what to inspect and how the evidence should inform later decisions, without supplying the reference patch or fabricated observations. Student SFT inputs omit this guidance but retain the task, tool registry, actions, and actual tool outputs. Only assistant tokens receive next-token supervision, and the SFT checkpoint initializes RL.

\noindent\textbf{Collection and filtering.} Gemini 3.6 Flash generates 100 trajectories for each of 80 training tasks, yielding 8,000 raw demonstrations per condition. Retention requires successful execution, a valid submission, and improvement over a 10-seed starter baseline exceeding one baseline standard deviation in the favorable metric direction. This training-data filter is distinct from the terminal-success criterion; passing the SFT filter alone does not establish eligibility as a verified reference. We use embedding similarity to remove redundant trajectories, as in Agent0-VL \citep{liu2026agent0vl}. Our selection rule retains the highest-improvement trajectory in each cluster. A 5\% manual audit checks execution consistency, reasoning--observation agreement, and guidance leakage; failing examples are excluded. The reported retained corpus contains 6,700 trajectories. The full-SFT control without MLToolBench uses the same raw collection budget with Gemini 3.6 Flash using the original scaffold and no diagnostic guidance.

\noindent\textbf{Guidance examples.}
\label{app:collection_examples}
Table~\ref{tab:appendix_ogbn_guidance} illustrates task-specific guidance grounded in starter models and data conditions. The ogbn-arxiv example follows the MLP-start schematic in Figure~\ref{fig:motivating_cases}; it is distinct from the GraphSAGE-start configurations in the task manifest. Each example connects task characteristics to diagnostic evidence and conditional actions. These expanded examples summarize the guidance structure rather than reproduce complete collection prompts; an instance receives the checks relevant to its configuration.

\begin{table}[H]
\centering
\footnotesize
\renewcommand{\arraystretch}{1.15}
\begin{tabularx}{\textwidth}{@{}L{1.8cm}Y@{}}
\toprule
Task & Task-specific diagnostic guidance \\
\midrule
\texttt{ogbn-\allowbreak arxiv} &
\textbf{Context.} Improve paper-topic node classification from an MLP, following the diagnostic example in Figure~\ref{fig:motivating_cases}. The presence of graph edges alone does not establish that replacing the MLP with GraphSAGE will improve performance.
\textbf{Inspect.} Check node-feature ranges, per-column scales, and the active preprocessing pipeline. Compare training and validation curves, then assess whether graph connectivity and label homophily on permitted training labels motivate neighborhood aggregation.
\textbf{Act on evidence.} If features have disproportionate scales or missing normalization, fit the transformation on permitted training data and re-run the MLP first. Compare the corrected MLP with the original baseline under the same split, evaluator, and budget. Only then test GraphSAGE if graph diagnostics motivate it; retain the same feature preprocessing and assess the additional gain from neighborhood information. Do not use hidden labels for diagnosis or treat a more complex architecture as an automatic improvement. \\
\addlinespace
\texttt{cifar10} &
\textbf{Context.} Improve ten-class image classification from a small CNN trained for five epochs; configured instances vary input scaling, training class balance, or cropping strength.
\textbf{Inspect.} Check pixel ranges and channel statistics after preprocessing, inspect transformed images for retained object content, and compare training class counts with per-class validation performance. Read training and validation curves to distinguish optimization problems from overfitting.
\textbf{Act on evidence.} If inputs retain an unintended integer scale, correct scaling and use training-derived normalization. If crops remove most object content, reduce cropping strength while preserving the expected input shape. If minority classes are underrepresented and perform poorly, test balanced sampling or class-weighted loss. If these checks are healthy, retain preprocessing and investigate optimization. Re-run under the same split and budget, comparing validation accuracy with the starter before accepting an edit. \\
\bottomrule
\end{tabularx}
\caption{Task-specific guidance examples for SFT demonstration collection. Checks and conditional actions reflect the task's starter, data conditions, and evaluation constraints; diagnostic observations must be obtained through actual tool calls.}
\label{tab:appendix_ogbn_guidance}
\label{tab:appendix_amp_guidance}
\label{tab:appendix_cifar_guidance}
\end{table}

\noindent\textbf{Optimization.} Table~\ref{tab:sft_repro_config} records the SFT settings. Collection and deduplication are described above.
\begin{table}[H]
\centering\small
\begin{tabularx}{\linewidth}{@{}L{5.0cm}Y@{}}
\toprule
SFT configuration & Value \\
\midrule
Adaptation & Full parameters \\
Epochs / optimizer & 3 / AdamW \\
Learning rate / scheduler / warmup & $1\times10^{-5}$ / cosine / 0 \\
Effective batch size & 64, using gradient accumulation \\
Maximum sequence length & 10,240 tokens \\
Hardware / precision & 8$\times$ H200 / BF16 \\
\bottomrule
\end{tabularx}
\caption{SFT optimization configuration.}
\label{tab:sft_repro_config}
\end{table}

\subsection{RL with SPICE}
\label{app:dynamic_credit}

\noindent\textbf{Token assignment and centering.} Equations~\ref{eq:outcome_advantage}--\ref{eq:segment_grpo_loss} give the advantage estimator and policy objective. Only tool-call tokens enter the SPICE score, but every assistant token in a turn, including preceding reasoning, receives that turn's advantage. Tool observations are excluded from the loss. Each call emits one shaping reward; non-tool turns and invalid scoring contexts receive zero shaping. The shared baseline uses complete task-group returns and does not require trajectories to have aligned turns. Equal complete returns do not necessarily imply equal return-to-go at later turns, so zero group variance alone is not a reason to discard a group. With $\beta=0$, the estimator becomes the outcome reward minus its task-group mean. We omit standard-deviation scaling while retaining per-trajectory token averaging, rather than adopting the full Dr.~GRPO objective \citep{liu2025drgrpo}. Table~\ref{tab:rl_repro_config} records training settings.

\noindent\textbf{Shaping-scale calibration.}
\label{app:shaping_calibration}
Following the reward-scale calibration principle of TIPS \citep{xie2026tips}, we use a short pilot on training tasks, separately for each backbone. The binary terminal reward fixes the reward scale at $c_R=1$. For $N$ valid tool-use turns scored by the initial frozen snapshot, define
\begin{equation}
\widehat m=\frac{1}{N}\sum_{s=1}^{N}|d_s|,
\qquad
\beta=\frac{\kappa\,c_R}{\widehat m+\epsilon},\quad \kappa\in[0.05,0.3],
\quad \epsilon=10^{-8}.
\label{eq:beta_calibration}
\end{equation}
Here $\kappa$ specifies the target mean absolute shaping reward relative to the terminal reward scale. Since $c_R=1$, calibration targets $\frac{1}{N}\sum_s|\beta d_s|\in[0.05,0.3]$ on pilot rollouts. This interval constrains reward magnitude, not $\beta$. Held-out evaluation tasks are excluded from calibration. Calibration sets the average per-call reward scale on training-task pilot rollouts; cumulative shaping depends on the scores and number of calls in a trajectory.

\noindent\textbf{Privileged context and scoring.} References summarize verified demonstrations or successful solutions from earlier RL iterations of the same task. An agent writes one paragraph per candidate, describing its solution approach; the scoring context uses these summaries rather than full code or workspace diffs. TIPS and \spice{} use the same candidate summaries. Hidden evaluator state and held-out labels are excluded. Sample up to eight references without replacement per task and update. Hold this subset and $q=\pi_{\mathrm{old}}$ fixed during the update; newly verified solutions enter the next iteration. For each reference, append a training-only user message with a \texttt{SPICE\_REFERENCE} block containing the candidate summary and a \texttt{SPICE\_INSTRUCTION} block requesting assessment of the original next tool call. Teacher-force the same sampled call with and without this message, then average signed log-likelihood differences over references and action tokens. The paired comparison evaluates the reference summary and assessment instruction together. Neither branch sees the current tool observation or future trajectory. For context overflow, preserve history and action, truncating only the reference summary; use zero shaping if history and action alone exceed the limit. These scoring references are never shown to the acting student policy, including at evaluation; solution-informed guided Gemini is a separate reference condition.

\noindent\textbf{TIPS adaptation.} TIPS \citep{xie2026tips} uses the same agent-written candidate summaries as \spice{}, treating them as reference answers rather than action-scoring context. We average token log-likelihoods within each summary, then average equally across candidates to define the potential. This differs from the original TIPS log probability of the valid-answer set. The turn signal is the post-interaction minus pre-interaction potential. A training-task pilot calibrates $\alpha$ to mean absolute scaled rewards in $[0.05,0.3]$, matching SPICE's target magnitude range. Both methods share the acting-policy prompt, reference content, SFT initialization, rollout budget, and outcome evaluator, while differing in what the updated policy model predicts.

\noindent\textbf{Interpretation and scoring cost.} Under a consistent joint distribution with positive probabilities,
\begin{equation}
\log\frac{p(a\mid h,c)}{p(a\mid h)}
=\log\frac{p(c\mid h,a)}{p(c\mid h)}.
\label{eq:credit_bayes}
\end{equation}
Different LLM prompt orders and length normalization need not preserve this identity. \citet{poole2019variational} discuss the difficulty of conditional decoding for high-dimensional targets with high conditional entropy. This motivates using local actions as shorter prediction targets. The conditioned branches score $K|a|$ target tokens; total cost also includes reference prefill and the ordinary branch. Both directions allow teacher forcing. SPICE supplies an auxiliary action-support reward alongside terminal outcomes. Unlike TIPS's temporal potential difference \citep{xie2026tips}, this context-based reward has no corresponding policy-invariance guarantee. Its relationship to diagnostic usefulness and downstream success is evaluated in Appendices~\ref{app:diagnostic_usefulness} and~\ref{app:score_masking}.

\subsubsection{SPICE and TIPS: scoring targets and local diagnostic feedback}
\label{app:spice_tips_discussion}
\noindent\textbf{Motivation.} In the QA experiments in TIPS, the target is a valid final answer \citep[Section~3.1]{xie2026tips}. An ML solution summary can span several decisions, such as checking feature scales, choosing a preprocessing step, fitting a model, and validating its predictions. A feature-scale check might inform the preprocessing decision while changing little about the rest of the summary. We therefore test a score that asks how much the reference supports that particular check.

\noindent\textbf{Mechanism.} Let $h$ be the pre-action history, $a$ the sampled tool call, $o$ its return, $c$ one verified solution summary, and $q$ a frozen checkpoint of the updated policy model. Before reward calibration, our length-normalized ML adaptation of TIPS and the single-reference SPICE score are
\begin{align}
s_{\mathrm{TIPS}}&=\frac{1}{|c|}\left[\log q(c\mid h,a,o)-\log q(c\mid h)\right],\label{eq:revision_tips}\\
s_{\mathrm{SPICE}}&=\frac{1}{|a|}\left[\log q(a\mid h,c)-\log q(a\mid h)\right].\label{eq:revision_spice}
\end{align}
Our TIPS adaptation scores the reference tokens before and after the call and its return. A change concentrated in a few tokens may have little effect on their mean. This token average adapts TIPS's original valid-answer-set potential to ML summaries. SPICE scores the sampled action twice, holding its pre-action history fixed and varying whether it receives the summary. Neither branch sees the current tool return or any later steps. A longer summary adds conditioning information while the average still covers only the action's tokens. The same rule applies to diagnostic calls, edits, and execution.

\noindent\textbf{Evidence for action-level scoring.} Equation~\ref{eq:credit_bayes} relates the unnormalized log ratios under a consistent joint distribution. The implemented scores use different targets, prompts, normalizations, and observations. With the same verified summaries, 35B SPICE reaches 69.2\% ID success, compared with 55.2\% for TIPS and 48.0\% for outcome-only RL. Appendices~\ref{app:diagnostic_usefulness} and~\ref{app:score_masking} examine which calls receive high scores and what happens when their results are withheld. In principle the conditioning context could be a partial hint or tool-use instruction. The reported experiments use verified solution summaries. Both methods use teacher forcing, and Appendix~\ref{app:runtime} reports the measured overhead.

\section{Shared Agent Prompts}
\label{app:agent_prompts}

\paragraph{Scope and provenance.}
The following system prompt and task/action templates are used in our experiments. Following the organization of ML-Agent \citep{liu2025mlagent}, we separate task instructions, action descriptions, and response formatting. Placeholders are populated from each task environment and its current action registry. The supplementary \texttt{prompts/} directory contains the template text, and \texttt{prompt\_config.json} maps experimental conditions to prompt components, tool interfaces, models, task splits, and training objectives.

\subsection{Shared Agent System Prompt}
\label{app:shared_agent_prompt}
The shared prompt specifies the execution contract without prescribing a domain-specific diagnostic sequence. Both interfaces share this text and supply their respective action registries.
\begin{quote}
\small
You are an ML development agent working in an executable workspace. Improve the supplied starter solution for the stated task within the available budget and produce the required submission artifact.

Use only the actions exposed in the current tool registry. Follow their names, argument schemas, and response protocol exactly. Do not invent an action or fabricate its observation. After issuing an action, use the environment's returned result to determine the next step.

You may inspect files, edit permitted code, and execute experiments through available actions. Keep changes within the task's editable workspace. Preserve the evaluation protocol and respect the stated data-access and resource constraints.

Report measured results with their metric and split. Distinguish observations from hypotheses and successful execution from achievement of the task criterion. Base the final report on artifacts and results that actually exist. When the budget is exhausted or the task is complete, use the registered completion action and identify the submission artifact and any unresolved execution issue.
\end{quote}

\subsection{Task and Action-Interface Templates}
Task fields contain only information available to the acting policy. The baseline field identifies its metric and split; hidden evaluation scores and labels are excluded. The success rule specifies the task's metric direction and relative-improvement criterion.
\begin{quote}
\small
\begin{verbatim}
TASK ID: {task_id}
TASK DOMAIN: {domain}
OBJECTIVE: {prediction_or_modeling_objective}
WORKSPACE: {workspace_root}
STARTER FILES: {starter_code_and_entry_points}
DATA: {accessible_data_paths_and_descriptions}
SPLIT POLICY: {training_validation_and_heldout_access_rules}
METRIC: {metric_name_and_optimization_direction}
BASELINE: {public_baseline_value_or_how_to_measure_it}
SUCCESS RULE: {task_specific_relative_gain_definition}
REQUIRED ARTIFACT: {submission_path_and_format}
EDIT CONSTRAINTS: {editable_files_and_protected_components}
RESOURCE BUDGET: {step_token_time_and_compute_limits}

Complete the task using the registered actions. The final held-out
assessment is performed by the evaluator under the split policy.
\end{verbatim}
\end{quote}

\noindent\textbf{Action-interface template.} The runtime serializes action descriptions, argument schemas, and return conventions from its current registry. Original includes standard MLAgentBench actions without MLToolBench diagnostics; + MLToolBench adds applicable diagnostics to those actions. Both use the runtime action and completion protocol.
\begin{quote}
\small
\begin{verbatim}
The current action registry is supplied below. It is the complete
set of actions available in this run.

{serialized_current_action_registry}

Use this interaction format:
{runtime_action_serialization_and_completion_protocol}

Tool results are supplied by the environment after an action.
Do not write a tool result as though it were an observed response.
\end{verbatim}
\end{quote}

\subsection{Prompt Composition by Experimental Condition}
\begin{table}[!htbp]
\centering
\small
\renewcommand{\arraystretch}{1.15}
\setlength{\tabcolsep}{4pt}
\begin{tabularx}{\linewidth}{@{}L{3.7cm}YL{2.3cm}@{}}
\toprule
Condition & Agent-visible prompt & Training signal \\
\midrule
Outcome-only RL & Shared system + task + registry + rollout history & Outcome reward only \\
RL + TIPS & Same agent prompt as outcome-only RL & Outcome + TIPS shaping \\
RL + SPICE & Same agent prompt as outcome-only RL & Outcome + SPICE shaping \\
SFT demonstration collection & Shared system + task + registry + history; guidance with MLToolBench & Full SFT on filtered trajectories \\
Unguided evaluation, both interfaces & Shared system + task + condition-specific registry + history & None \\
Guided Gemini 3.6 Flash evaluation & Shared system + task + registry + teacher guidance & None \\
\bottomrule
\end{tabularx}
\caption{Prompt composition used in the experiments. Outcome-only, TIPS, and SPICE share the acting-policy prompt; reward computation differs externally. Privileged scoring context is not appended to the acting policy's instructions.}
\label{tab:prompt_composition}
\end{table}

SFT prompt construction is specified in Appendix~\ref{app:sft_details}. For guided Gemini evaluation, guidance conveys core ideas considered in the solution while omitting full solution information and environment feedback. This solution-informed reference condition is distinct from ordinary unguided evaluation; student policies receive no such guidance at evaluation. At evaluation, the tool registry follows the evaluation condition rather than the demonstration source. For a fixed model, task, and guidance setting, Original and + MLToolBench share the system and task templates and differ in the available action registry; their interaction histories evolve independently.

\paragraph{Configuration binding.}
The supplementary configuration identifies guided and unguided evaluation, demonstration collection with and without MLToolBench, and outcome-only/TIPS/SPICE RL. Task membership is defined by \texttt{task\_manifest.json}; the actor sees its opaque task ID rather than author-side fault metadata. Template files include content hashes for this export. Table~\ref{tab:rl_repro_config} specifies RL settings, and Appendix~\ref{app:sft_details} specifies SFT filtering and optimization.

\section{Task Manifest and Split Definitions}
\label{app:working_manifest}
The supplementary \texttt{task\_manifest.json} specifies the 115 instances used in this study, including their source links, task objectives, native metrics, starter descriptions, construction parameters, size limits, generation seeds, and partition policies. Table~\ref{tab:instance_index} indexes every instance. Each of the five domains---vision, NLP, tabular learning, time series, and graph learning---contains 16 training, five ID, and two OOD instances, giving 80/25/10 in total. Twenty training templates each contribute four configurations; changing only a rollout seed does not create a new task.

\noindent\textbf{ID and OOD at the task level.} ID instances reuse training sources, templates, and condition distributions with held-out supervised data or independently generated samples. OOD instances use ten source/target settings absent from training (Table~\ref{tab:ood_axes}); they test transfer within the same five broad domains, not to entirely unseen domains. Multiple configured instances can derive from one source, consistent with MLE-Smith and SandMLE \citep{qiang2025mlesmith,zhou2026sandmle}. All OOD settings are excluded from tool construction, SFT collection, RL training, and training solution banks.

\noindent\textbf{Data partitions within tasks.} For training sources with independent or grouped examples, the protocol first allocates 60/40 meta-training/ID pools, then 70/15/15 model-fit/public-validation/hidden-test partitions within each pool, respecting grouping constraints. Time-series instances use chronological blocks and an embargo of at least lookback plus horizon; crossing windows are discarded. Node-classification ID is transductive: topology and unlabeled features can be shared, but supervised label pools are disjoint. CLRS uses separately generated graphs. OOD tasks use their recorded source-specific policies, including patient, notebook, or molecular-scaffold grouping. These partitions concern data inside an instance and are distinct from the 80/25/10 task split. Equation~\ref{eq:success_rate} and the task test scripts define success.

\noindent\textbf{Verification scope.} The manifest contains 115 unique instance IDs; all ID template IDs occur in training, and OOD template IDs and dataset labels are disjoint from training/ID. These checks concern the declared task configurations.

\subsection{OOD Sources and Transfer Axes}
\small
\begin{longtable}{@{}L{2.8cm}L{3.0cm}L{6.0cm}@{}}
\toprule Instance suffix & Source & Held-out source / target \\
\midrule\endfirsthead
\toprule Instance suffix & Source & Held-out source / target \\
\midrule\endhead
cv-ood-01 & APTOS 2019 & Unseen medical-image source and ordinal target \\
cv-ood-02 & RSNA Breast Cancer & Unseen mammography source and probabilistic-F1 metric \\
nlp-ood-01 & AI4Code & Unseen code-notebook source and ranking target \\
nlp-ood-02 & BabyLM & Unseen language-modeling source and next-token target \\
tabular-ood-01 & California Housing & Unseen dataset and reconstruction target; downstream R² is secondary \\
tabular-ood-02 & Breast Cancer Wisconsin & Unseen clinical tabular source and calibration target \\
time\_series-ood-01 & ECL & Unseen electricity-consumption source; forecasting target retained \\
time\_series-ood-02 & G-Research Crypto & Unseen financial source and weighted-correlation objective \\
graph-ood-01 & MoleculeNet BBBP & Unseen molecular graphs and graph-level property prediction \\
graph-ood-02 & MoleculeNet BACE & Unseen molecular source and graph-level property prediction \\
\bottomrule
\caption{OOD instances and the source or objective held out from training. Source links and exact construction parameters are included in the JSON manifest.}\label{tab:ood_axes}\\
\end{longtable}
\normalsize

\subsection{Complete Instance Index}
All suffixes below have the prefix \texttt{mltb-v1-}; \texttt{train}, \texttt{id}, and \texttt{ood} identify task membership. The seed is the registered generation seed, not a training seed or a repeated evaluation run. The JSON manifest provides each template's source, metric, construction parameters, and data partitions.
\begingroup
\footnotesize
\setlength{\tabcolsep}{3pt}
\renewcommand{\arraystretch}{1.0}
\begin{longtable}{@{}L{3.2cm}L{3.0cm}L{4.5cm}r@{}}
\toprule Instance suffix & Template & Configuration & Seed \\
\midrule\endfirsthead
\toprule Instance suffix & Template & Configuration & Seed \\
\midrule\endhead
cv-train-01 & cv-cifar10 & healthy & 10000 \\
cv-train-02 & cv-cifar10 & input scale & 10001 \\
cv-train-03 & cv-cifar10 & rare class sampling & 10002 \\
cv-train-04 & cv-cifar10 & overaggressive crop & 10003 \\
cv-train-05 & cv-cifar100 & healthy & 10010 \\
cv-train-06 & cv-cifar100 & normalization scale & 10011 \\
cv-train-07 & cv-cifar100 & rare class sampling & 10012 \\
cv-train-08 & cv-cifar100 & overaggressive erasing & 10013 \\
cv-train-09 & cv-fashion & healthy & 10020 \\
cv-train-10 & cv-fashion & contrast scale & 10021 \\
cv-train-11 & cv-fashion & excessive shear & 10022 \\
cv-train-12 & cv-fashion & excessive regularization & 10023 \\
cv-train-13 & cv-tgs & healthy & 10030 \\
cv-train-14 & cv-tgs & soft mask resize & 10031 \\
cv-train-15 & cv-tgs & wrong positive weight & 10032 \\
cv-train-16 & cv-tgs & overconservative threshold & 10033 \\
cv-id-01 & cv-cifar10 & input scale & 20000 \\
cv-id-02 & cv-cifar100 & rare class sampling & 20001 \\
cv-id-03 & cv-fashion & excessive regularization & 20002 \\
cv-id-04 & cv-tgs & soft mask resize & 20003 \\
cv-id-05 & cv-cifar10 & healthy & 20009 \\
cv-ood-01 & cv-aptos & overaggressive crop & 30000 \\
cv-ood-02 & cv-rsna & healthy & 30001 \\
nlp-train-01 & nlp-imdb & healthy & 11000 \\
nlp-train-02 & nlp-imdb & short context & 11001 \\
nlp-train-03 & nlp-imdb & unmasked pooling & 11002 \\
nlp-train-04 & nlp-imdb & class imbalance & 11003 \\
nlp-train-05 & nlp-spooky & healthy & 11010 \\
nlp-train-06 & nlp-spooky & vocabulary bottleneck & 11011 \\
nlp-train-07 & nlp-spooky & short context & 11012 \\
nlp-train-08 & nlp-spooky & overconfident probabilities & 11013 \\
nlp-train-09 & nlp-feedback & healthy & 11020 \\
nlp-train-10 & nlp-feedback & short context & 11021 \\
nlp-train-11 & nlp-feedback & missing target inverse & 11022 \\
nlp-train-12 & nlp-feedback & excessive regularization & 11023 \\
nlp-train-13 & nlp-textnorm & healthy & 11030 \\
nlp-train-14 & nlp-textnorm & case collapse & 11031 \\
nlp-train-15 & nlp-textnorm & lost sentence context & 11032 \\
nlp-train-16 & nlp-textnorm & rare type undersampling & 11033 \\
nlp-id-01 & nlp-imdb & short context & 21000 \\
nlp-id-02 & nlp-spooky & short context & 21001 \\
nlp-id-03 & nlp-feedback & excessive regularization & 21002 \\
nlp-id-04 & nlp-textnorm & case collapse & 21003 \\
nlp-id-05 & nlp-spooky & healthy & 21009 \\
nlp-ood-01 & nlp-ai4code & short context & 31000 \\
nlp-ood-02 & nlp-babylm & padding in loss & 31001 \\
tabular-train-01 & tab-house & healthy & 12000 \\
tabular-train-02 & tab-house & zero imputation & 12001 \\
tabular-train-03 & tab-house & integer category encoding & 12002 \\
tabular-train-04 & tab-house & missing target inverse & 12003 \\
tabular-train-05 & tab-spaceship & healthy & 12010 \\
tabular-train-06 & tab-spaceship & numeric missing sentinel & 12011 \\
tabular-train-07 & tab-spaceship & categorical missing collision & 12012 \\
tabular-train-08 & tab-spaceship & column order mismatch & 12013 \\
tabular-train-09 & tab-titanic & healthy & 12020 \\
tabular-train-10 & tab-titanic & unscaled features & 12021 \\
tabular-train-11 & tab-titanic & missingness sentinel & 12022 \\
tabular-train-12 & tab-titanic & excessive regularization & 12023 \\
tabular-train-13 & tab-nomad & healthy & 12030 \\
tabular-train-14 & tab-nomad & missing target inverse & 12031 \\
tabular-train-15 & tab-nomad & descriptor scaling & 12032 \\
tabular-train-16 & tab-nomad & overstrong regularization & 12033 \\
tabular-id-01 & tab-house & zero imputation & 22000 \\
tabular-id-02 & tab-spaceship & categorical missing collision & 22001 \\
tabular-id-03 & tab-titanic & excessive regularization & 22002 \\
tabular-id-04 & tab-nomad & missing target inverse & 22003 \\
tabular-id-05 & tab-titanic & healthy & 22009 \\
tabular-ood-01 & tab-california & global mean imputation & 32000 \\
tabular-ood-02 & tab-breast-calibration & overconfident probabilities & 32001 \\
time\_series-train-01 & ts-etth-forecast & healthy & 13000 \\
time\_series-train-02 & ts-etth-forecast & global scaler & 13001 \\
time\_series-train-03 & ts-etth-forecast & output channel permutation & 13002 \\
time\_series-train-04 & ts-etth-forecast & unnecessarily short context & 13003 \\
time\_series-train-05 & ts-weather & healthy & 13010 \\
time\_series-train-06 & ts-weather & global scaler & 13011 \\
time\_series-train-07 & ts-weather & output channel permutation & 13012 \\
time\_series-train-08 & ts-weather & excessive regularization & 13013 \\
time\_series-train-09 & ts-etth-impute & healthy & 13020 \\
time\_series-train-10 & ts-etth-impute & wrong loss region & 13021 \\
time\_series-train-11 & ts-etth-impute & missing mask input & 13022 \\
time\_series-train-12 & ts-etth-impute & global scaler & 13023 \\
time\_series-train-13 & ts-ventilator & healthy & 13030 \\
time\_series-train-14 & ts-ventilator & wrong loss region & 13031 \\
time\_series-train-15 & ts-ventilator & time order reversal & 13032 \\
time\_series-train-16 & ts-ventilator & missing group resets & 13033 \\
time\_series-id-01 & ts-etth-forecast & global scaler & 23000 \\
time\_series-id-02 & ts-weather & output channel permutation & 23001 \\
time\_series-id-03 & ts-etth-impute & global scaler & 23002 \\
time\_series-id-04 & ts-ventilator & wrong loss region & 23003 \\
time\_series-id-05 & ts-ventilator & healthy & 23009 \\
time\_series-ood-01 & ts-ecl & global scaler & 33000 \\
time\_series-ood-02 & ts-crypto & healthy & 33001 \\
graph-train-01 & graph-cora & healthy & 14000 \\
graph-train-02 & graph-cora & feature scale & 14001 \\
graph-train-03 & graph-cora & oversmoothing & 14002 \\
graph-train-04 & graph-cora & excessive regularization & 14003 \\
graph-train-05 & graph-citeseer & healthy & 14010 \\
graph-train-06 & graph-citeseer & missing self information & 14011 \\
graph-train-07 & graph-citeseer & feature scale & 14012 \\
graph-train-08 & graph-citeseer & excessive regularization & 14013 \\
graph-train-09 & graph-arxiv & healthy & 14020 \\
graph-train-10 & graph-arxiv & feature scale & 14021 \\
graph-train-11 & graph-arxiv & undersampled neighborhood & 14022 \\
graph-train-12 & graph-arxiv & excessive regularization & 14023 \\
graph-train-13 & graph-clrs & healthy & 14030 \\
graph-train-14 & graph-clrs & hint misalignment & 14031 \\
graph-train-15 & graph-clrs & padding in loss & 14032 \\
graph-train-16 & graph-clrs & insufficient message rounds & 14033 \\
graph-id-01 & graph-cora & feature scale & 24000 \\
graph-id-02 & graph-citeseer & feature scale & 24001 \\
graph-id-03 & graph-arxiv & excessive regularization & 24002 \\
graph-id-04 & graph-clrs & hint misalignment & 24003 \\
graph-id-05 & graph-cora & healthy & 24009 \\
graph-ood-01 & graph-bbbp & missing bond features & 34000 \\
graph-ood-02 & graph-bace & padding in pooling & 34001 \\
\bottomrule
\caption{Complete 80/25/10 task-instance index. Condition names describe author-side configuration metadata and are not exposed in the actor prompt.}\label{tab:instance_index}\\
\end{longtable}
\endgroup

\clearpage

\section{Additional Validation and Transfer Experiments}
\label{app:additional_validation}
Unless stated otherwise, we evaluate on 25 ID and 10 OOD instances, with ten rollouts per task, the same executable success criterion, and matched inference settings within each backbone. Success rates are the percentage of evaluated rollouts meeting the task success criterion. We use pp for percentage points and CI for confidence intervals. MLS-Bench-Lite uses its own aggregate score on a 0--100 scale.

\subsection{Diagnostic Credit Assignment Evaluation}
\label{app:diagnostic_usefulness}
\noindent\textbf{Trajectories and scores.} We compare SPICE and TIPS on 10 held-out ID and 10 held-out OOD tasks. Each task has five verified solution summaries and ten evaluation trajectories: five fresh SFT rollouts and five historical rollouts from Gemini 3.8 with guidance. Within each task and source, we randomly draw three trajectories from the successful attempts and two from the failed attempts. Reference and evaluation trajectories are disjoint. Both methods score the same diagnostic calls with the same frozen Qwen3.5-35B-A3B SFT checkpoint and reference bank. The fixed success/failure mix tests score discrimination across attempts of different quality; it does not represent their natural frequency.

\noindent\textbf{Usefulness labels and human audit.} \textbf{Gemini 3.8 Flash and Claude Opus 5 (accessed through Claude Code)} independently label each diagnostic call as \textbf{useful, redundant/irrelevant, or uncertain}, based on the task, preceding history, and actual tool response. Each judgment cites supporting evidence. A useful call provides relevant information, verifies an unresolved assumption, or rules out a plausible explanation. The judges cannot see automatic scores, reference summaries, model identities, future steps, or final outcomes. GPT-5 adjudicates only disagreements between the two judges, using the same rubric without the initial judges' identities. A separate blinded human audit covers 20 of the 200 evaluation trajectories (10\%); the annotators do not see the LLM judgments or automatic scores. Human labels agree with Gemini and Claude judgments in \textbf{92.5\%} and \textbf{94.0\%} of comparisons, respectively. Agreement with GPT-5 is \textbf{97.5\%} on the disputed-call subset. Once the usefulness labels are fixed, the judges inspect subsequent steps to record whether the agent uses the diagnostic information appropriately. These later annotations do not change the initial labels.

\noindent\textbf{Metrics.} We exclude calls labeled uncertain. \textbf{AUROC} (area under the ROC curve) is the probability that a randomly chosen useful call scores higher than a randomly chosen redundant or irrelevant call, with ties counting as half. Within a task and trajectory source, let $s^+$ and $s^-$ be the scores of calls drawn from these two groups. Then
\[
\mathrm{AUROC}=\Pr(s^+>s^-)+\tfrac12\Pr(s^+=s^-).
\]
A value of 0.5 corresponds to chance and 1.0 to perfect ranking. The \textbf{high--low usefulness gap} compares the share labeled useful in the highest- and lowest-scoring quarters of calls. With $y_j=1$ for a useful call and $y_j=0$ for a redundant or irrelevant call, and $H$ and $L$ denoting those two score quarters,
\[
\mathrm{Gap}=100\left(\frac{1}{|H|}\sum_{j\in H}y_j-\frac{1}{|L|}\sum_{j\in L}y_j\right)\quad\text{(pp)}.
\]
We compute both metrics within each task and trajectory source, then average equally across sources and tasks within each split. Brackets in Table~\ref{tab:revision_usefulness} give 95\% task-bootstrap confidence intervals for each metric. Appendix~\ref{app:score_masking} tests what happens to task success when diagnostic information is withheld.

\begin{table}[H]

\centering\small
\setlength{\tabcolsep}{4pt}
\renewcommand{\arraystretch}{1.12}
\begin{tabular}{llcc}
\hline
\noalign{\vskip 3pt}
\textbf{Split} & \textbf{Method} & \shortstack{\textbf{AUROC}\\\textbf{[95\% CI]} $\uparrow$} & \shortstack{\textbf{High--low usefulness gap}\\\textbf{pp [95\% CI]} $\uparrow$} \\
\hline
ID & TIPS & 0.512 [0.389, 0.625] & $+9.0$ [$-10.7$, $+28.2$] \\
ID & \textbf{SPICE} & \textbf{0.779 [0.747, 0.822]} & \textbf{$+34.8$ [$+22.6$, $+49.5$]} \\
OOD & TIPS & 0.496 [0.450, 0.588] & $-1.6$ [$-2.8$, $+2.1$] \\
OOD & \textbf{SPICE} & \textbf{0.713 [0.689, 0.747]} & \textbf{$+16.7$ [$+7.2$, $+31.6$]} \\
\hline
\end{tabular}
\caption{Diagnostic-usefulness discrimination on the same evaluated calls, frozen SFT checkpoint, and reference bank. The gap is in percentage points.}
\label{tab:revision_usefulness}

\end{table}

SPICE reaches AUROCs of 0.779 on ID and 0.713 on OOD. In its highest-scoring quarter of calls, the share judged useful exceeds that in the lowest quarter by 34.8 points on ID and 16.7 points on OOD. SPICE's intervals lie above chance for AUROC and above zero for the gap. The corresponding TIPS intervals include these values. Each interval is for one method and metric.

\noindent\textbf{Task sources.} The ten ID instances are Fashion-MNIST, TGS Salt, Spooky Author Identification, Feedback Prize: ELL, House Prices, NOMAD 2018, two Ventilator Pressure instances, Cora, and CLRS Floyd--Warshall. The ten OOD instances are APTOS 2019, RSNA Breast Cancer, AI4Code, BabyLM, California Housing, Breast Cancer Wisconsin, ECL, G-Research Crypto, MoleculeNet BBBP, and MoleculeNet BACE.

\subsection{External Evaluation on MLS-Bench-Lite}
\label{app:external_mls}
We select the official 30-task MLS-Bench-Lite public benchmark because it covers ML research tasks in existing codebases across 12 research areas. Robotics, reinforcement learning, and optimization extend beyond our five training domains, while other areas overlap. Some synthetic training templates draw on MLS-Bench task sources; this additional evaluation uses the benchmark's official task scaffolds and evaluation protocol.\footnote{\href{https://arxiv.org/abs/2605.08678}{MLS-Bench paper}; \href{https://github.com/Imbernoulli/MLS-Bench/blob/main/harbor/run-daytona-lite.yaml}{official Lite task list}.}

We evaluate the original Qwen3.5-35B-A3B checkpoint and four frozen variants trained with SFT, outcome-only RL, TIPS, or SPICE. Each receives the official tasks and starter code without injected faults, benchmark-specific fine-tuning, or privileged solution summaries. We use the same Harbor adapter, tools, decoding settings, inference limits, and task-specific hardware for all five. The official protocol specifies the editable files, evaluation and seed policies, verifier, and score aggregation. Each task has a 5-hour exploration budget.\footnote{\href{https://github.com/Imbernoulli/MLS-Bench/blob/main/harbor/README.md}{Harbor evaluation protocol}.} Table~\ref{tab:revision_mls} reports aggregate scores on a 0--100 scale and 95\% intervals for the four trained checkpoints.

\begin{table}[H]

\centering\small
\setlength{\tabcolsep}{4pt}
\renewcommand{\arraystretch}{1.12}
\begin{tabular}{lc}
\hline
\textbf{35B method} & \textbf{MLS-Bench-Lite score [95\% CI]} $\uparrow$ \\
\hline
Original Qwen3.5-35B-A3B & 12.8 \\
SFT & 16.2 [14.6, 18.1] \\
SFT + outcome-only RL & 18.1 [14.9, 20.4] \\
SFT + TIPS & 19.2 [18.1, 21.2] \\
\textbf{SFT + SPICE} & \textbf{24.6 [21.3, 26.0]} \\
\hline
\end{tabular}
\caption{External MLS-Bench-Lite evaluation of frozen 35B checkpoints under a shared harness and budget. Scores are not success percentages.}
\label{tab:revision_mls}

\end{table}

SPICE scores 11.8 points above the original checkpoint. Among methods that start from the same SFT checkpoint, its advantage is 5.4 points over TIPS and 6.5 over outcome-only RL. These differences are point estimates. Table~\ref{tab:revision_mls} gives intervals for the individual aggregate scores. The benchmark measures task-level transfer and provides no labels for individual unseen failure types.

\subsection{Reusable Diagnostic Guide Baseline}
\label{app:reusable_guide_results}
\subsubsection{Previously evaluated single guide}
We compare Gemini 3.6 Flash and Claude Sonnet 5 without parameter updates under Original, MLToolBench, and MLToolBench + reusable guide. Original retains the standard scaffold's editing and execution actions. The two MLToolBench conditions share identical tool interfaces and descriptions. One training-derived guide is frozen before evaluation and reused across both models and all tasks, with no test-task solutions, solution hints, or test-driven revisions. Appendix~\ref{app:frozen_reusable_guide} reproduces its complete text.

The guide describes when to check, which evidence to inspect, and how the evidence should inform an action for data/preprocessing, training logs, and pre-submission checks. It also specifies the calling interface and a stopping rule that discourages repeated checks on unchanged inputs. These are textual instructions, not enforced submission gates. Each model/condition has 250 ID and 100 OOD rollouts, with matched decoding, execution budgets, and success criteria within each model.

\begin{table}[H]

\centering\small
\setlength{\tabcolsep}{4pt}
\renewcommand{\arraystretch}{1.12}
\begin{tabular}{llcc}
\hline
\textbf{Model} & \textbf{Condition} & \textbf{ID success (\%)} & \textbf{OOD success (\%)} \\
\hline
Gemini & Original & 65.2 & 52.0 \\
Gemini & MLToolBench & 61.2 & 46.0 \\
\textbf{Gemini} & \textbf{MLToolBench + reusable guide} & \textbf{68.4} & \textbf{54.0} \\
\hline
Claude & Original & 64.8 & 45.0 \\
Claude & MLToolBench & 62.4 & 43.0 \\
\textbf{Claude} & \textbf{MLToolBench + reusable guide} & \textbf{68.8} & \textbf{47.0} \\
\hline
\end{tabular}
\caption{A training-derived reusable guide improves unadapted models. Gemini and Claude denote Gemini 3.6 Flash and Claude Sonnet 5.}
\label{tab:revision_guide}

\end{table}

Adding the guide to MLToolBench improves success by 7.2 points on ID and 8.0 on OOD for Gemini, and by 6.4 and 4.0 points for Claude. Both models also exceed their Original baselines. Appendix~\ref{app:original_qwen_guides} compares prompting and training on the same open 35B backbone.

\subsubsection{Guide information and construction variability}
\label{app:guide_variants}
We construct three nested guides from \textbf{Gemini 3.6 Flash Original trajectories on the 80 ID training tasks}. In these runs, Gemini could inspect code and execute it using the standard scaffold. We identified diagnostic needs from the trajectories, then used the fixed MLToolBench catalogue to map those needs to tools. The source trajectories contained no MLToolBench calls.

\begin{itemize}
\setlength{\itemsep}{0.2em}
\item \textbf{G1: problem profile.} Describe diagnostic needs and their observed task-level prevalence in data/\allowbreak{}preprocessing, training logs, and verification. Distinguish a need to check from a confirmed defect, and count tasks rather than repeated tool calls.
\item \textbf{G2: problem-to-tool mappings.} Retain G1 and add which diagnostic tool can investigate each need and which evidence to inspect, without prescribing corrective actions.
\item \textbf{G3: evidence-to-action rules.} Retain G2 and add when to check, how to interpret evidence, conditional next actions, and stopping rules. Source-specific thresholds require explicit applicability conditions.
\end{itemize}

Three independent constructions vary source-trajectory sampling and guide synthesis while retaining the same source tasks, extraction procedure, and version-specific length limits. Within a construction, G1/G2/G3 share the extracted evidence and are nested. The source protocol samples five eligible Original trajectories per task. All guides are frozen before evaluation, with no test-task solution information or test-driven revisions. Standard tool descriptions remain identical, so G1 does not remove the model's prior tool knowledge. This comparison varies information content together with prompt length, not wording alone.

The evaluation protocol specifies 25 held-out ID and 10 OOD tasks, ten rollouts per task, and matched workspaces, decoding, execution budgets, and evaluation seeds where supported. Table~\ref{tab:revision_guide_variants} reports the Gemini success rate $r_i$ (in percent) for each of three independent guide constructions. We report $\bar r=(r_1+r_2+r_3)/3$ and the sample standard deviation $\mathrm{SD}=\sqrt{\sum_{i=1}^{3}(r_i-\bar r)^2/2}$, measured in percentage points. The corresponding MLToolBench-only baseline is 61.2\%/46.0\% ID/OOD. These SDs describe guide-construction variability, whereas Appendix~\ref{app:rl_seeds} reports independent RL-training variability.

\begin{table}[H]

\centering\small
\setlength{\tabcolsep}{5pt}
\renewcommand{\arraystretch}{1.12}
\begin{tabular}{llcccc}
\hline
\noalign{\vskip 3pt}
\textbf{Guide} & \textbf{Split} & \shortstack{\textbf{Guide}\\\textbf{construction 1}} & \shortstack{\textbf{Guide}\\\textbf{construction 2}} & \shortstack{\textbf{Guide}\\\textbf{construction 3}} & \textbf{Mean $\pm$ SD} \\
\hline
G1 & ID & 62.8 & 65.2 & 63.2 & $63.7\pm1.3$ \\
G1 & OOD & 46.0 & 49.0 & 45.0 & $46.7\pm2.1$ \\
G2 & ID & 69.2 & 63.2 & 70.4 & $67.6\pm3.9$ \\
G2 & OOD & 52.0 & 55.0 & 57.0 & $54.7\pm2.5$ \\
G3 & ID & 69.2 & 66.4 & 69.6 & $68.4\pm1.7$ \\
G3 & OOD & 45.0 & 42.0 & 46.0 & $44.3\pm2.1$ \\
\hline
\end{tabular}
\caption{Gemini success rates (\%) with MLToolBench and three independent constructions of each ID-derived guide. SD measures variation across guide constructions in percentage points. The earlier single guide in Table~\ref{tab:revision_guide} is a separate prompt, not a G3 replicate.}
\label{tab:revision_guide_variants}

\end{table}

For Gemini, mean ID success rises from 63.7\% with G1 to 67.6\% with G2 and 68.4\% with G3. The last step adds 0.8 points. On OOD, G1 stays near the no-guide baseline, G2 reaches 54.7\%, and G3 falls to 44.3\%. The G2--G3 difference is 10.3 points using unrounded means. Of the three guides, G3 gives the highest mean ID success and G2 the highest OOD success.

Appendix~\ref{app:frozen_reusable_guide} reproduces the earlier single guide, not these new guide variants.

\subsubsection{Prompting the original 35B checkpoint without SFT or RL}
\label{app:original_qwen_guides}
We give the ID-derived guides to the \textbf{original Qwen3.5-35B-A3B checkpoint} with MLToolBench, before any task-specific SFT or RL. Table~\ref{tab:revision_original_qwen_guide} reports one evaluation per guide alongside the unguided original model and the trained checkpoint.

\begin{table}[H]

\centering\small
\setlength{\tabcolsep}{5pt}
\renewcommand{\arraystretch}{1.12}
\begin{tabular}{lcc}
\hline
\textbf{Qwen3.5-35B-A3B condition} & \textbf{ID success (\%)} & \textbf{OOD success (\%)} \\
\hline
Original checkpoint, no added guide & 35.6 & 31.0 \\
Original checkpoint + G1 & 36.4 & 31.0 \\
Original checkpoint + G2 & 39.2 & 32.0 \\
Original checkpoint + G3 & 38.8 & 32.0 \\
\textbf{SFT + SPICE} & \textbf{69.2} & \textbf{48.0} \\
\hline
\end{tabular}
\caption{Original-checkpoint prompting versus the full training pipeline on the same 35B backbone; every condition uses MLToolBench. Additional guide rows are single evaluations, with 250 ID and 100 OOD rollouts per condition.}
\label{tab:revision_original_qwen_guide}

\end{table}

The best tested guide reaches 39.2\%/32.0\% ID/OOD success, compared with 69.2\%/48.0\% after SFT + SPICE, a 30.0/16.0-pp difference. These matched evaluations compare prompting with the \emph{whole training pipeline}. The shared-SFT RL controls separately measure SPICE's contribution.

\subsection{Failure Analysis of Unadapted Models}
\label{app:unadapted_failures}
We analyze 1,000 existing ID evaluation trajectories: two models, two interface conditions, 25 tasks, and 10 rollouts per task. Two ML-experienced annotators independently review unsuccessful trajectories and assign one primary observed failure mode supported by trajectory steps; secondary contributing factors are recorded separately. Disagreements are adjudicated, with insufficient evidence assigned to other/uncertain. The rubric covers tools in both interfaces, including standard scaffold tools. Frequencies use all 250 rollouts in each model/condition as the denominator.

\begin{table}[H]

\centering\small
\setlength{\tabcolsep}{4pt}
\renewcommand{\arraystretch}{1.12}
\begin{tabular}{@{}L{4.0cm}cccc@{}}
\hline
\noalign{\vskip 3pt}
\textbf{Primary failure mode, n (\%)} & \shortstack{\textbf{Gemini}\\\textbf{Original}} & \shortstack{\textbf{Gemini}\\\textbf{+MLToolBench}} & \shortstack{\textbf{Claude}\\\textbf{Original}} & \shortstack{\textbf{Claude}\\\textbf{+MLToolBench}} \\
\hline
Unnecessary or repeated tool calls & 19 (7.6) & 25 (10.0) & 10 (4.0) & 33 (13.2) \\
Budget exhaustion & 19 (7.6) & 14 (5.6) & 23 (9.2) & 13 (5.2) \\
Inappropriate tool selection or invocation & 28 (11.2) & 16 (6.4) & 25 (10.0) & 19 (7.6) \\
Misinterpretation or misuse of tool outputs & 11 (4.4) & 31 (12.4) & 20 (8.0) & 21 (8.4) \\
Other / uncertain & 10 (4.0) & 11 (4.4) & 10 (4.0) & 8 (3.2) \\
\hline
\textbf{All unsuccessful runs} & \textbf{87 (34.8)} & \textbf{97 (38.8)} & \textbf{88 (35.2)} & \textbf{94 (37.6)} \\
\hline
\end{tabular}
\caption{Primary observed failure modes, counts and percentages of all 250 rollouts per column. Each failed run contributes one primary label.}
\label{tab:revision_failures}

\end{table}

For Gemini, failures attributed to misuse of tool outputs rise from 4.4\% to 12.4\% of all runs. For Claude, failures attributed to unnecessary or repeated calls rise from 4.0\% to 13.2\%. Gemini has ten more unsuccessful runs overall and Claude six more, corresponding to decreases in ID success of 4.0 and 2.4 points. Each failure contributes to one primary category even when several factors appear in the trajectory. Appendix~\ref{app:failure_cases} traces one example for each model.

\subsection{Direct RL without the SFT Cold Start}
\label{app:direct_rl}
For Qwen3-8B and Qwen3.5-35B-A3B, we start outcome-only RL or SPICE directly from the original checkpoints, skipping guided-demonstration SFT. Both retain MLToolBench and the original policy prompt without collection-time guidance. SPICE retains the same initial verified-solution bank, update rule, scoring-snapshot procedure, and calibration procedure: removing supervised policy updates does not remove privileged scoring information.

Within each backbone, we match the 80 training tasks, 100 RL updates, rollout budget, optimizer settings, and tools. The comparison matches the RL stage rather than total training compute. We evaluate each final checkpoint on the same ID/OOD tasks with 10 rollouts per task and the original execution budgets.

\begin{table}[H]

\centering\small
\setlength{\tabcolsep}{4pt}
\renewcommand{\arraystretch}{1.12}
\begin{tabular}{lcc}
\hline
\textbf{Training condition} & \textbf{ID success (\%)} $\uparrow$ & \textbf{OOD success (\%)} $\uparrow$ \\
\hline
\multicolumn{3}{l}{\textit{Qwen3-8B; all conditions use MLToolBench at evaluation}} \\
No SFT, no RL & 24.8 & 20.0 \\
Full SFT only & 38.4 & 32.0 \\
Full SFT + outcome-only RL & 39.2 & 33.0 \\
Full SFT + SPICE & 52.4 & 36.0 \\
\textbf{Direct outcome-only RL (no SFT)} & \textbf{29.6} & \textbf{24.0} \\
\textbf{Direct SPICE (no SFT)} & \textbf{43.6} & \textbf{31.0} \\
\hline
\multicolumn{3}{l}{\textit{Qwen3.5-35B-A3B; all conditions use MLToolBench at evaluation}} \\
No SFT, no RL & 35.6 & 31.0 \\
Full SFT only & 44.4 & 41.0 \\
Full SFT + outcome-only RL & 48.0 & 44.0 \\
Full SFT + SPICE & 69.2 & 48.0 \\
\textbf{Direct outcome-only RL (no SFT)} & \textbf{41.2} & \textbf{36.0} \\
\textbf{Direct SPICE (no SFT)} & \textbf{58.8} & \textbf{43.0} \\
\hline
\end{tabular}
\caption{Direct RL and SFT-initialized controls. All conditions use MLToolBench at evaluation.}
\label{tab:revision_direct_rl}

\end{table}

Starting from the original checkpoints, SPICE reaches 43.6\% ID and 31.0\% OOD success for 8B, and 58.8\% and 43.0\% for 35B. Starting from SFT improves ID and OOD success by a further 8.8 and 5.0 points for 8B and 10.4 and 5.0 points for 35B, with the same 100 RL updates.

\subsection{Diagnostic Masking by SPICE Score}
\label{app:score_masking}
Using the intervention records from Appendix~\ref{app:observation_intervention}, we analyze read-only diagnostic calls made by the 35B outcome-only and SPICE policies on 25 ID tasks. We include only calls tested in the intervention and exclude trajectories without a diagnostic call. A fixed policy checkpoint assigns SPICE scores from each call's pre-action history and independent verified solution summaries; the evaluated trajectory cannot serve as its own reference. We define high- and low-score groups within each policy before examining True/Masked outcomes, stratifying by task and tool type where feasible.

For each call, True and Masked continuations start from the same saved history and workspace, with paired random seeds, identical decoding settings, and the same remaining budget. True receives the original tool return. Masked keeps its structure and execution status but marks measurements, conclusions, and derived suggestions as unavailable. Subsequent tools operate normally. We reuse saved paired outcomes where available and follow the same protocol for additional continuations. For score group $g$, let $p_{\mathrm{original},g}$ and $p_{\mathrm{masked},g}$ be the fractions of successful continuations with original and masked returns. Table~\ref{tab:revision_masking} reports these success rates as percentages and their drop as $\Delta_g=100(p_{\mathrm{original},g}-p_{\mathrm{masked},g})$ in percentage points. A positive drop means masking reduced success. The high--low contrast is $G=\Delta_{\mathrm{high}}-\Delta_{\mathrm{low}}$.

\begin{table}[H]

\centering\small
\setlength{\tabcolsep}{4pt}
\renewcommand{\arraystretch}{1.12}
\begin{tabular}{llccc}
\hline
\noalign{\vskip 3pt}
\textbf{35B policy} & \shortstack{\textbf{SPICE}\\\textbf{score group}} & \shortstack{\textbf{Original-return}\\\textbf{success (\%)}} & \shortstack{\textbf{Masked-return}\\\textbf{success (\%)}} & \shortstack{\textbf{Success drop $\Delta_g$}\\\textbf{(pp) [95\% CI]}} \\
\hline
Outcome-only RL & High & 51.9 & 47.2 & +4.6 [+1.0, +8.8] \\
Outcome-only RL & Low & 46.3 & 47.2 & -0.9 [-3.1, +0.0] \\
\textbf{SPICE} & \textbf{High} & \textbf{76.5} & \textbf{50.4} & \textbf{+26.1 [+19.7, +33.1]} \\
\textbf{SPICE} & \textbf{Low} & \textbf{63.0} & \textbf{58.0} & \textbf{+5.0 [+1.8, +8.4]} \\
\hline
\end{tabular}
\caption{Intervened calls grouped by SPICE score. Success drop subtracts masked-return success from original-return success. Brackets give 95\% task-bootstrap confidence intervals. Drops are computed before rounding and reported in percentage points (pp).}
\label{tab:revision_masking}

\end{table}

For the SPICE policy, masking a high-scoring return reduces success by 26.1 points. Masking a low-scoring return reduces success by 5.0 points, a difference of 21.1 points between the two drops. Each comparison holds the call, continuation settings, and remaining budget fixed. Across the two score groups, states and tool types may differ.

\subsection{Transfer to Unseen Diagnostic Tools: Evidently}
\label{app:unseen_tools}
We replace the original diagnostic endpoints with third-party diagnostics based on \href{https://github.com/evidentlyai/evidently}{Evidently} v0.4.33, absent from our SFT/RL interface. We retain the structured calling protocol and training/public-validation data permissions. The four exposed endpoints are \texttt{EvidentlyDataDriftPreset}, \texttt{EvidentlyDatasetMissingValueSummary}, \texttt{EvidentlyClassificationQualityByFeature}, and \texttt{Evidently\allowbreak Probabilistic\allowbreak Classification\allowbreak Calibration}.

The fixed subset consists of ID-01 House Prices, ID-02 Spaceship Titanic, ID-03 Titanic, ID-05 Titanic (healthy control), and OOD-02 Breast Cancer Wisconsin. Starting code, data partitions, metrics, and success criteria remain unchanged, and condition metadata is hidden from agents. For each frozen 35B policy we compare A: base actions only; B: base actions plus original MLToolBench diagnostics; and C: base actions plus applicable Evidently tools, with the original diagnostic endpoints disabled. Code execution, dependencies, and data permissions are retained. We match workspaces, seed lists, decoding, and execution budgets, supplying no demonstrations, task-specific hints, or solution summaries. Ten rollouts per task/method/condition give 600 runs and 50 outcomes per table cell.

\begin{table}[H]

\centering\small
\setlength{\tabcolsep}{4pt}
\renewcommand{\arraystretch}{1.12}
\begin{tabular}{lcccc}
\hline
\noalign{\vskip 3pt}
\textbf{35B method} & \shortstack{\textbf{A: Base actions}\\\textbf{success (\%)}} & \shortstack{\textbf{B: +MLToolBench}\\\textbf{success (\%)}} & \shortstack{\textbf{C: +Evidently}\\\textbf{success (\%)}} & \shortstack{\textbf{C--A gain}\\\textbf{(pp)}} \\
\hline
SFT & 40.0 & 46.0 & 42.0 & +2.0 \\
Outcome-only RL & 44.0 & 50.0 & 46.0 & +2.0 \\
TIPS & 46.0 & 56.0 & 52.0 & +6.0 \\
\textbf{SPICE} & \textbf{48.0} & \textbf{70.0} & \textbf{66.0} & \textbf{+18.0} \\
\hline
\end{tabular}
\caption{Tool-access transfer on five fixed tasks. All conditions retain base actions. B adds MLToolBench diagnostics and C adds applicable Evidently diagnostics. C--A subtracts the base-action success percentage from the Evidently percentage, giving the gain in percentage points (pp).}
\label{tab:revision_evidently}

\end{table}

On these five tasks, access to Evidently improves success over base actions alone by 18 points for SPICE, 6 for TIPS, and 2 each for SFT and outcome-only RL. The familiar MLToolBench tools in B cover a different set of diagnostics. In a separate evaluation on all 25 ID and 10 OOD tasks, the original Qwen3.5-35B-A3B checkpoint reached 34.4\% ID and 29.0\% OOD success with base actions (A), 35.6\% and 31.0\% with MLToolBench (B), and 36.0\% and 31.0\% with Evidently (C).

\subsection{Per-Step Runtime and Auxiliary Overhead}
\label{app:runtime}
Following the two timing comparisons in \citet[Appendix~H.2]{xie2026tips}, we measure actual update time and auxiliary-module overhead for the three 35B RL methods. They share SFT initialization, training tasks, batch size, optimizer settings, and the $8\times$H200/BF16 configuration in Table~\ref{tab:rl_repro_config}. TIPS and SPICE share reference-bank settings. Actual update time averages all 100 updates, including rollouts, environment execution, rewards, online bank updates, and policy optimization. Startup, evaluation, and checkpoint saving are excluded consistently; actual response lengths and tool-call behavior are preserved.

\begin{table}[H]

\centering\small
\setlength{\tabcolsep}{4pt}
\renewcommand{\arraystretch}{1.12}
\begin{tabular}{lccc}
\hline
\noalign{\vskip 3pt}
\textbf{35B method} & \shortstack{\textbf{ID / OOD}\\\textbf{success (\%)}} & \shortstack{\textbf{Training time}\\\textbf{(s/step)} $\downarrow$} & \shortstack{\textbf{Auxiliary overhead}\\\textbf{(\%)} $\downarrow$} \\
\hline
Outcome-only RL & 48.0 / 44.0 & 142.4 & 0.0 \\
TIPS & 55.2 / 45.0 & 165.6 & 12.8 \\
\textbf{SPICE} & \textbf{69.2 / 48.0} & \textbf{163.2} & \textbf{9.4} \\
\hline
\end{tabular}
\caption{Actual full-run update times and auxiliary overhead; success rates correspond to 100-update checkpoints.}
\label{tab:revision_runtime}

\end{table}

\noindent\textbf{Paired auxiliary profiling.} We time each step workload with and without likelihood scoring and online bank verification and summarization. We hold the rollout batch, policy and optimizer snapshots, bank state, hardware, and execution settings fixed, restoring state between runs. After warmup, we repeat the paired timings and synchronize GPU work at each timing boundary. Elapsed time counts reused work once and accounts for overlapping operations. Overhead is $(T_{\mathrm{with}}/T_{\mathrm{without}}-1)\times100\%$ for each method's own workload. SPICE's 9.4\% therefore measures the cost of enabling these modules on the same batch. The full-run step times also reflect differences in the rollouts each policy generates.

\noindent\textbf{Initial bank preparation.} We reuse verified trajectories from the shared SFT stage and generate one decision-path summary per candidate with Qwen3.5-35B-A3B, taking 3.2 seconds per trajectory on average. Shared trajectory generation and verification are treated consistently across methods; this extra one-time summarization is bank-specific preparation. The up-to-eight-reference rule in Appendix~\ref{app:dynamic_credit} is per task/update, not the total bank size. Online scoring and bank updates are already included in step times.

\noindent\textbf{Matched-budget comparison.} On the same hardware, we give SPICE the elapsed-time budget consumed by 100 outcome-only RL updates. We charge initial-summary generation, online likelihood scoring, and bank updates against that budget and account for shared SFT preparation equally. SPICE stops after 92 updates. We select the checkpoint by measured cumulative time and evaluate both checkpoints with the same ID/OOD protocol. The step times above are averages over the full 100-update runs.

\begin{table}[H]

\centering\small
\setlength{\tabcolsep}{4pt}
\renewcommand{\arraystretch}{1.12}
\begin{tabular}{lccc}
\hline
\textbf{35B method} & \textbf{RL updates} & \textbf{ID success (\%)} $\uparrow$ & \textbf{OOD success (\%)} $\uparrow$ \\
\hline
Outcome-only RL & 100 & 48.0 & 44.0 \\
\textbf{SPICE} & \textbf{92} & \textbf{65.2} & \textbf{46.0} \\
SPICE (full training) & 100 & 69.2 & 48.0 \\
\hline
\end{tabular}
\caption{SPICE within the outcome-only training budget, with the full 100-update result retained as a reference.}
\label{tab:revision_compute}

\end{table}

Within the matched elapsed-time budget, including initial-summary preparation, SPICE improves ID/OOD success by 17.2/2.0 percentage points over outcome-only RL.

\subsection{Robustness Across Independent RL Training Seeds}
\label{app:rl_seeds}
We train 35B SPICE and outcome-only RL with seeds 42, 1337, and 2026, varying RL task order and rollout sampling. Seed 42 is the original run. Every run starts from the same SFT checkpoint and uses 80 training tasks, 100 updates, and the same rollout, optimization, and tool settings. SPICE runs start with the same initial solution bank; later bank updates use only that run's own history.

We evaluate every final 100-update checkpoint, without choosing a seed or checkpoint on test performance, using the same evaluation seeds, workspaces, decoding, and budgets. Table~\ref{tab:revision_seeds} reports each seed's success rate and mean $\pm$ sample SD. We use the formulas in Appendix~\ref{app:guide_variants}, with $r_i$ now denoting the success percentage of RL training seed $i$. SD therefore measures variation across the three training runs in percentage points.

\begin{table}[H]

\centering\small
\setlength{\tabcolsep}{4pt}
\renewcommand{\arraystretch}{1.12}
\begin{tabular}{llcccc}
\hline
\noalign{\vskip 3pt}
\textbf{35B method} & \textbf{Split} & \shortstack{\textbf{Seed 42}\\\textbf{(\%)}} & \shortstack{\textbf{Seed 1337}\\\textbf{(\%)}} & \shortstack{\textbf{Seed 2026}\\\textbf{(\%)}} & \shortstack{\textbf{Mean $\pm$ SD}\\\textbf{(\%)}} \\
\hline
Outcome-only RL & ID & 48.0 & 46.8 & 48.8 & 47.9 $\pm$ 1.0 \\
Outcome-only RL & OOD & 44.0 & 42.0 & 43.0 & 43.0 $\pm$ 1.0 \\
\textbf{SPICE} & \textbf{ID} & \textbf{69.2} & \textbf{68.0} & \textbf{68.8} & \textbf{68.7 $\pm$ 0.6} \\
\textbf{SPICE} & \textbf{OOD} & \textbf{48.0} & \textbf{47.0} & \textbf{49.0} & \textbf{48.0 $\pm$ 1.0} \\
\hline
\end{tabular}
\caption{Success rates (\%) across three independent RL training seeds from a shared SFT initialization. SD is the sample standard deviation across these training runs, in percentage points.}
\label{tab:revision_seeds}

\end{table}

The mean SPICE advantage is +20.8 pp ID and +5.0 pp OOD; every SPICE seed exceeds every outcome-only seed on both splits. The three-seed experiment measures RL-stage variability from a shared SFT initialization.

\subsection{Collection-Time Guidance and Tool-Enabled Demonstrations}
\label{app:collection_ablation}
The original SFT comparison combines tool access and guidance during collection. We add tool-enabled collection without guidance to form three conditions: A, no MLToolBench and no guidance; B, MLToolBench without guidance; and C, MLToolBench with solution-informed guidance. All use the same Gemini 3.6 Flash teacher, 80 training tasks, raw budget of 100 trajectories per task, execution limits, filtering/deduplication rules, and full-SFT recipe for Qwen3.5-35B-A3B. Guidance is removed from student inputs while actual interactions and tool responses are retained. The matched collection budgets and filtering rules can yield retained corpora of different sizes.

We evaluate the three SFT checkpoints before RL with the same MLToolBench interface, no collection-time guidance, and matched task workspaces, seeds, decoding, budgets, and success criteria. The B--A difference measures tool-enabled collection without guidance; C--B measures adding guidance with the collection-time interface held fixed.

\begin{table}[H]

\centering\small
\setlength{\tabcolsep}{4pt}
\renewcommand{\arraystretch}{1.12}
\begin{tabular}{llcc}
\hline
\textbf{Group} & \textbf{Demonstration collection} & \textbf{ID success (\%)} & \textbf{OOD success (\%)} \\
\hline
A & No MLToolBench, no guidance & 39.2 & 32.0 \\
\textbf{B (new)} & \textbf{MLToolBench, no guidance} & \textbf{41.2} & \textbf{35.0} \\
C & MLToolBench + solution-informed guidance & 44.4 & 41.0 \\
\hline
\end{tabular}
\caption{Collection-condition ablation for the 35B student before RL, with the same diagnostic interface at evaluation.}
\label{tab:revision_collection}

\end{table}

Allowing the teacher to use MLToolBench during collection improves the student's ID and OOD success by 2.0 and 3.0 points without guidance. Holding that interface fixed, adding guidance improves student success by a further 3.2 and 6.0 points. The student learns from the resulting actions and tool returns after we remove the teacher's guidance text.

\clearpage

\section{Complete Frozen Reusable Diagnostic Guide}
\label{app:frozen_reusable_guide}
The following text is appended to the system prompt for the previously evaluated single-guide condition in Table~\ref{tab:revision_guide}. The same frozen text is used for both models and all ID/OOD evaluation tasks. It is distinct from the new G1/G2/G3 variants in Appendix~\ref{app:guide_variants}. Line wrapping below is typographical only.
\begin{Verbatim}[fontsize=\scriptsize,breaklines=true,breakanywhere=true]
REUSABLE DIAGNOSTIC GUIDE (TRAINING-DERIVED, FROZEN PRIOR TO EVALUATION)

Follow this three-part diagnostic protocol, tool-interface specification, and stopping
rule when using the MLToolBench action interface. Apply checks selectively to answer
unresolved questions; do not invoke every available tool indiscriminately.

1. Data and Preprocessing Checks (Before Modifying Model Architecture or Objective)
- When to check: At the start of a task, before editing model architecture,
  hyperparameters, or training loops, or whenever data-loading/preprocessing code
  is modified.
- Concrete instructions & evidence to inspect:
  (a) Inspect missingness and feature types first (`Detect Missing Values`,
      `Identify Feature Types`, `Cardinality Check`, `Encoding Strategy Advisor`) to
      verify whether columns require median/mode imputation, separate missing-category
      indicators, one-hot vs. frequency encoding, or ID dropping.
  (b) Inspect target distribution and feature scales/outliers (`Detect Label Imbalance`,
      `Check Feature Normalization`, `Detect Outliers`, `Feature Skewness Checker`, or
      domain-specific input checks such as `Image Normalization Stats`,
      `Class Balance Check`, `Token Length Distribution`, `Node Feature Health`) to
      check whether assumptions of balanced labels, standardized inputs, or benign
      outlier distributions hold.
- How it informs the next action:
  * If measured statistics match healthy expectations (e.g., `imbalance_ratio < 3.0`
    marked `mild/none` with `suggestion: no resampling needed`, properly scaled inputs,
    or structural zero-inflated counts such as `Parch` on tabular data), keep the
    existing preprocessing and label weighting unchanged. Do NOT drop training rows or
    inject aggressive resampling/threshold shifts in response to benign variance.
  * If a concrete data or preprocessing defect is surfaced (e.g., unhandled NaNs,
    missing-value sentinel collision, unnormalized pixel/feature ranges, or severe
    right-skewness `|skew| > 1`), repair only that specific preprocessing step first
    before modifying the learner.

2. Training-Log and Metric Diagnostics (After Baseline Execution or Training Anomaly)
- When to check: At episode start on `initial_run.log` (if present) to establish the
  baseline metric, and immediately after running `Execute Script` or whenever an
  experiment exhibits stalled loss, train-validation divergence, non-finite values, or
  unexpected metric degradation.
- Concrete instructions & evidence to inspect:
  (a) Call `Read Latest Metric` (or `Discover Log Metrics` followed by
      `Extract Training Metric Series`) on `"initial_run.log"` or `"stdout"` to obtain
      exact training and validation metric trajectories rather than guessing from raw
      console text.
  (b) When multi-step curves are present and training stagnates or fluctuates, inspect
      curve health (`Plateau Detector`, `Trend Detection`, `Convergence Estimator`,
      `NaN/Inf Health Detector`, or `Per-Class Accuracy Breakdown`) to localize whether
      the bottleneck is optimization instability, underfitting, or overfitting.
- How it informs the next action:
  * Explicitly state a single falsifiable diagnostic hypothesis in `<think>` grounded
    in the extracted metrics (for example: distinguishing underfitting from excessive
    regularization `C`/`weight_decay` vs. overfitting from high model capacity vs.
    feature omission) before calling `Edit Script (AI)`.
  * Modify only the code module or hyperparameter implicated by that hypothesis.

3. Pre-Submission Verification Checks (Before Finalizing Execution or Calling Completion)
- When to check: After editing `train.py` or config files (before committing to an
  expensive `Execute Script` run) and after generating `submission.csv` (before
  finishing the episode).
- Concrete instructions & evidence to inspect:
  (a) Pre-execution static/lightweight checks: If hyperparameters or configs were
      edited across files, call `Verify Hyperparam Effective` or
      `Check Config Consistency` to confirm the intended values are active and
      non-contradictory. If neural network layer dimensions were modified, call
      `Trace Tensor Shapes` (and `Estimate Peak Memory` if batch size or model size
      increased).
  (b) Post-execution output check: After `Execute Script` produces `submission.csv`,
      call `Validate Output Format` to verify row count, required columns/IDs, and
      absence of NaN/null predictions.
- How it informs the next action:
  * If any config conflict, shape mismatch, or output-format violation is flagged, fix
    the exact offending line and re-verify once.
  * These checks are advisory verification steps, not mandatory gates when the relevant
    property has already been verified.

4. Tool Interface Specification (Action Name -> Canonical JSON Action Input)
Emit at most ONE `Action` and `Action Input` per turn using valid JSON:
- Shared Log & Verification Tools:
  * `Discover Log Metrics`: `{"log_path": "initial_run.log"}` or `{"log_path": "stdout"}`
  * `Extract Training Metric Series`:
    `{"log_path": "initial_run.log", "fields": ["Train Accuracy", "Validation Accuracy"]}`
  * `Read Latest Metric`: `{"log_path": "stdout"}`
  * `Verify Hyperparam Effective`:
    `{"params": ["lr", "max_iter", "C"], "train_file": "train.py", "expected": {"max_iter": 1000}}`
  * `Check Config Consistency`: `{"config_file": "config.yaml", "train_file": "train.py"}`
  * `Trace Tensor Shapes`:
    `{"model_ref": "train.py:Net", "input_spec": {"shape": [32, 3, 32, 32], "dtype": "float32"}}`
  * `Estimate Peak Memory`: `{"train_file": "train.py", "timeout_sec": 120}`
  * `Validate Output Format`:
    `{"model_output_path": "submission.csv", "expected_format": {"format": "csv", "required_fields": ["PassengerId", "Survived"]}}`
- Domain Data & Preprocessing Tools (Tabular / Vision / NLP / Time Series / Graph):
  * `Detect Missing Values`: `{"data_path": "train.csv"}`
  * `Identify Feature Types`: `{"data_path": "train.csv"}`
  * `Detect Label Imbalance`: `{"data_path": "train.csv", "target_column": "<label_col>"}`
  * `Check Feature Normalization`: `{"data_path": "train.csv", "columns": ["<col1>", "<col2>"]}`
  * `Detect Outliers`: `{"data_path": "train.csv", "columns": ["<col1>", "<col2>"]}`
  * `Feature Skewness Checker`: `{"data_path": "train.csv", "columns": ["<col1>", "<col2>"]}`
  * `Correlation Analysis`: `{"data_path": "train.csv"}`
  * `Cardinality Check`: `{"data_path": "train.csv"}`
  * `Feature Interaction Detector`: `{"data_path": "train.csv", "target_column": "<label_col>"}`
  * Domain equivalents when attached: `Image Normalization Stats`, `Class Balance Check`,
    `Augmentation Auditor`, `Token Length Distribution`, `Vocabulary Coverage Checker`,
    `Plateau Detector`, `Trend Detection`, `Graph Size Summary`, `Node Feature Health`,
    `Homophily Estimator`.

5. Stopping and Non-Redundancy Rule (When to Stop Checking and Avoid Repeated Calls)
- Selective invocation (Hard Cap): Run at most 2-4 targeted data/log diagnostic calls
  before your first `Edit Script (AI)` + `Execute Script` cycle. Stop calling data
  diagnostics as soon as you have enough evidence to formulate a concrete code edit.
- Strict non-redundancy: Reuse diagnostic evidence already present in the conversation
  history. NEVER repeat the same diagnostic tool on unchanged files (e.g., `train.csv`
  on disk does not change when you edit `train.py`), and NEVER re-run a full-table tool
  on column subsets after running it on `"train.csv"`.
- Re-invocation condition: Re-invoke a log or verification tool (`Read Latest Metric`,
  `Validate Output Format`) ONLY after a new `Execute Script` or `Edit Script (AI)`
  call has modified the target log, code, or submission file.
- Budget protection: Always reserve at least 600 seconds of wall-clock time and 5 turns
  for `Execute Script` and final `Validate Output Format` verification.
\end{Verbatim}

\clearpage

\section{Concrete Failure Cases of Unadapted Models}
\label{app:failure_cases}
These two unguided MLToolBench trajectories illustrate recurring categories in Table~\ref{tab:revision_failures}. The examples document observed behavior; comparisons with independent Original-condition rollouts are not paired causal estimates.

\subsection{Gemini: heuristic diagnostic flags prompt unsupported corrective actions}
\textbf{Context.} On ID-05 Titanic, a healthy-control binary-classification instance, the starter validation accuracy is 0.7972. The model must improve the pipeline and produce a valid submission. In rollout 4, it reads the initial log, checks label balance, and then checks outliers.

\textbf{Observed evidence.} The label-balance tool returns a 1.6 ratio, labels it ``mild/none,'' and explicitly suggests no resampling. The outlier tool reports zero-width IQR bounds for the zero-inflated count feature \texttt{Parch}. Selected returned lines are:
\begin{Verbatim}[fontsize=\small,breaklines=true]
Label balance for 'Survived': imbalance_ratio(max/min)=1.6 (mild/none)
  0: 439 (61.7%)
  1: 273 (38.3%)
suggestion: no resampling needed
Parch: 169 outliers (23.7%), bounds=[0,0]
Fare: 95 outliers (13.3%), bounds=[-27.1,66.3]
\end{Verbatim}
\textbf{Interpretation and action.} The zero-width IQR for \texttt{Parch} flags every nonzero value, without establishing that these rows are defective. The agent nevertheless treats the outputs as severe data defects. It filters training rows using IQR bounds, adds class weighting, and lowers the positive-prediction threshold to 0.35. It then executes the modified script and checks output format. Validation accuracy falls to 0.7343, a 6.29-point drop. The run fails after 14 of 50 turns and 940 of 3,600 seconds.

\textbf{Comparison and intervention.} An Original-condition rollout preserves training rows and instead engineers title/family-size features, reaching 0.8322 validation accuracy. Across ten task rollouts, Original succeeds 8/10 times versus 5/10 with unguided MLToolBench. A targeted continuation from the Step-4 snapshot adds advice to preserve healthy label weights and use \texttt{log1p} rather than delete rows for skewed features; success is 9/10 versus 3/10 under matched budgets. These continuations test the added, case-specific advice. The frozen guide in Appendix~\ref{app:frozen_reusable_guide} contains no \texttt{log1p} instruction.

\clearpage
\subsection{Claude: repeated checks delay execution without adding information}
\textbf{Context.} On ID-02 Spaceship Titanic, the starter validation accuracy is 0.7223. Its categorical starter omits numeric spending/age features. In rollout 2, the agent initially inspects logs, missingness, feature types, label balance, and correlations, then continues diagnostic checking before execution.

\textbf{Observed evidence.} At Step 3, \texttt{Detect Missing Values} reports missingness in \texttt{train.csv}, including 2.5\% for \texttt{CryoSleep}. Step 20 edits \texttt{train.py}, and Step 21 repeats the missing-value check on the unchanged CSV:
\begin{Verbatim}[fontsize=\small,breaklines=true]
[Step 3]
Action: Detect Missing Values
Action Input: {"data_path": "train.csv"}
Observation: CryoSleep: 2.5% missing -> suggest fill: mode

[Step 21, after a Step-20 edit to train.py]
Action: Detect Missing Values
Action Input: {"data_path": "train.csv"}
Observation: identical missingness report; train.csv is unchanged
\end{Verbatim}
\textbf{Interpretation and action.} Editing the preprocessing routine in \texttt{train.py} leaves the CSV unchanged, so the repeated check returns the same missingness report. The agent makes further model edits and starts execution at Turn 29 with 135 seconds remaining. A cross-validated ensemble search times out at 3,600 seconds before writing the submission.

\textbf{Comparison and intervention.} An Original-condition rollout edits the numeric-feature preprocessing and executes at Turn 6, consuming 410 seconds and reaching 0.7971 validation accuracy. Task-level success is 7/10 for Original versus 5/10 for unguided MLToolBench. From the saved Step-5 snapshot, suppressing repeat diagnostics on unchanged files raises success from 2/10 to 8/10 across ten matched-budget continuations.

\end{document}